\documentclass[times,review,preprint,authoryear]{elsarticle}

 \usepackage{medima}
\usepackage{framed,multirow}
\usepackage{amsmath}
\usepackage{amssymb}
\usepackage{subfigure}
\usepackage{booktabs}
\usepackage{dcolumn}
\usepackage{array}
\usepackage{color}
\usepackage{caption}   
\usepackage{subcaption}
\usepackage{float}    
\usepackage{algorithm}
\usepackage{algpseudocode}
\usepackage[T1]{fontenc}
\usepackage{amssymb}
\usepackage{latexsym}

\usepackage{url}
\usepackage{xcolor}

\usepackage{hyperref}

\definecolor{newcolor}{rgb}{.8,.349,.1}
\definecolor{customgreen}{RGB}{58,124,34}
\definecolor{customred}{RGB}{153,26,26}

\begin{document}

\verso{Chenyu Zhang \textit{et~al.}}

\begin{frontmatter}

\title{Relational Anatomical Supervision for Accurate 3D Multi-Chamber Cardiac Mesh Reconstruction}%

\author[1]{Chenyu \snm{Zhang}}

\author[1]{Yihao \snm{Luo}}
\author[5,6]{Lei \snm{Zhu}}
\author[1]{Martyn G \snm{Boutelle}}
\author[1]{Choon Hwai \snm{Yap}}
\author[1,2,3,4]{Guang \snm{Yang}\corref{cor1}}
\cortext[cor1]{Corresponding author}
\ead{g.yang@imperial.ac.uk}
\address[1]{Bioengineering Department Imperial College London, London W12 7SL, United Kingdom}
\address[2]{National Heart and Lung Institute, Imperial College London, London, United Kingdom}
\address [3]{Cardiovascular Research Centre, Royal Brompton Hospital, London SW3 6NP, United Kingdom; }
\address[4]{School of Biomedical Engineering \& Imaging Sciences, King's College London, London WC2R 2LS, United Kingdom}
\address[5]{ROAS Thrust, Hong Kong University of Science and Technology (Guangzhou), Guangzhou, China}
\address[6]{Department of Electronic and Computer Engineering, Hong Kong University of Science and Technology, Hong Kong SAR, China}


\begin{abstract}
Accurate reconstruction of multi-chamber cardiac anatomy from medical images is a cornerstone for patient-specific modeling, physiological simulation, and interventional planning. However, current reconstruction pipelines fundamentally rely on surface-wise geometric supervision and model each chamber in isolation, resulting in anatomically implausible inter-chamber violations despite apparently favorable overlap or distance metrics. In this work, we propose a relational anatomical supervision framework for multi-chamber cardiac mesh reconstruction by introducing a Mesh Interrelation Enhancement (MIE) loss. The proposed formulation explicitly encodes spatial relationships between cardiac structures into a differentiable occupancy-based objective, thereby transforming qualitative anatomical rules into quantitative geometric supervision. We further establish violation-aware evaluation metrics to directly quantify inter-chamber structural correctness, revealing systematic limitations of commonly used geometric measures such as Dice and Chamfer distance. Extensive experiments on multi-center CT data, densely sampled MR data, and two independent external cohorts, including a highly heterogeneous congenital heart disease population, demonstrate that the proposed method consistently suppresses clinically critical boundary violations by up to 83\%, while maintaining competitive volumetric accuracy and achieving superior surface fidelity. Notably, the proposed relational supervision generalizes robustly across imaging modalities, centers, and pathological conditions, even under severe anatomical deformation. These results demonstrate that distance-based supervision alone is insufficient to guarantee anatomically faithful reconstruction, and that explicit enforcement of multi-structure anatomical relations provides a principled and robust pathway toward reliable patient-specific cardiac modeling.
\end{abstract}


\end{frontmatter}


\section{Introduction}
\begin{figure*}[!t] \centerline{\includegraphics[width=\columnwidth]{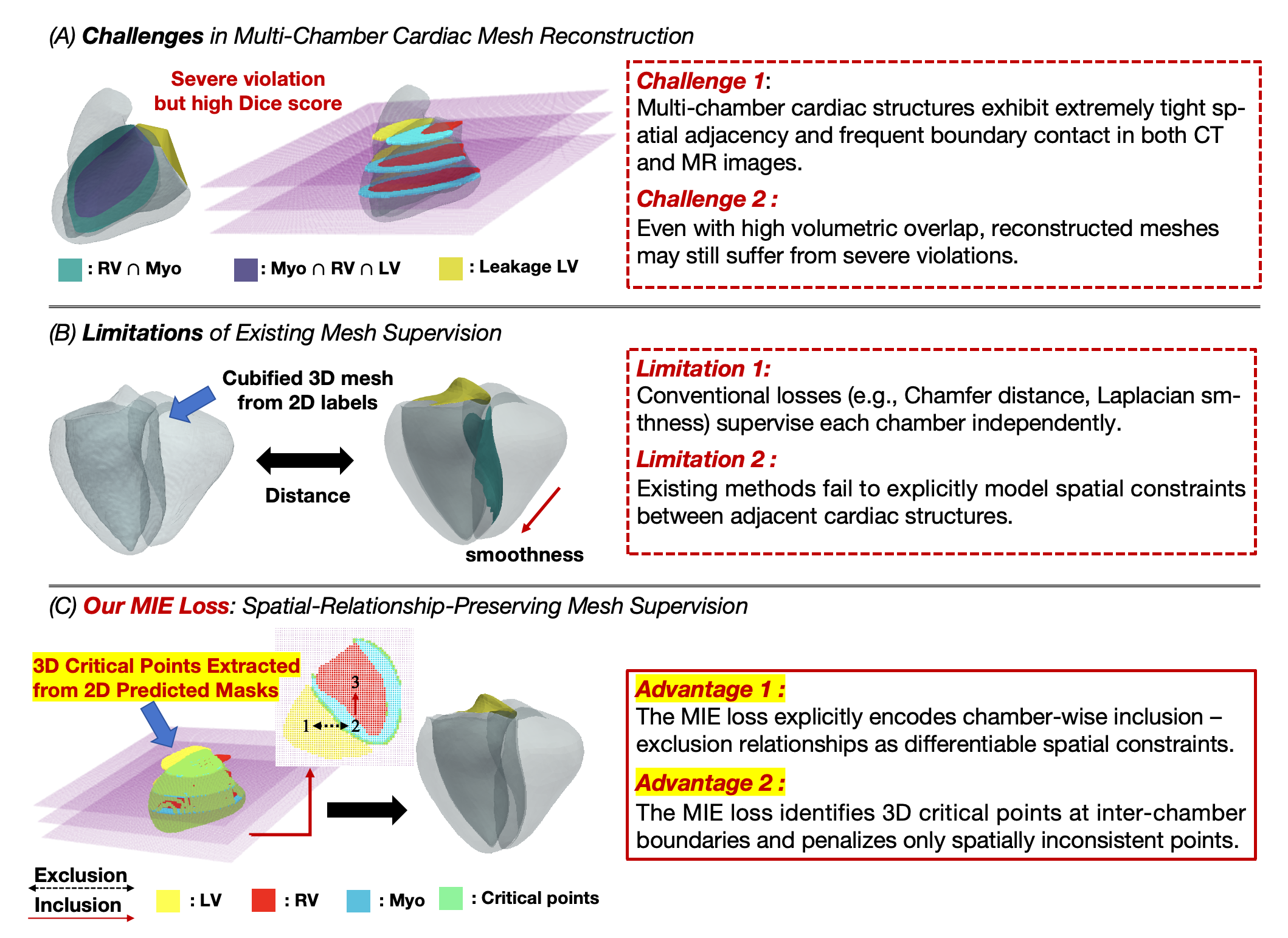}} \caption{ Motivation of spatial-relationship–preserving mesh supervision for multi-chamber cardiac reconstruction. (A) Challenges in Multi-Chamber Cardiac Mesh Reconstruction. (B) Limitations of Existing Mesh Supervision. (C) Our MIE Loss: Spatial-Relationship-Preserving Mesh Supervision
.} 
\label{fig1} 
\end{figure*}

Accurate and anatomically consistent 3D cardiac mesh reconstruction is vital for faithfully simulating cardiac function, from tissue mechanics \cite{marx2020personalization} and electrophysiology \cite{trayanova2011electromechanical} to personalized interventions\cite{trayanova2020personalized,sel2024building}. Yet, conventional deformation-based or segmentation-driven pipelines are prone to anatomically inconsistent structural errors, including unnatural inter-chamber connections and inter-structure leakage~\cite{kong2021whole}, as shown in Fig.~\ref{fig1}. These structural violations not only undermine physiological plausibility but also compromise downstream tasks, including hemodynamic simulations \cite{mittal2016computational} and treatment planning \cite{kong2022learning}.

While deep learning has been widely applied to reconstruct cardiac structures from volumetric data \cite{kong2021deep,augustin2016anatomically,maher2019accelerating}, most approaches do not explicitly enforce consistency between anatomical components. Voxel-level constraints~\cite{acebes2024centerline,shit2021cldice} can incorporate certain spatial priors at the volumetric level, but they do not directly operate in the mesh domain or enforce surface-level geometric relations. Meanwhile, common mesh regularizers such as Chamfer distance~\cite{fan2017point} constrain geometric surface proximity, while Laplacian smoothing~\cite{rustamov2007laplace} promotes local surface regularity; however, neither explicitly prevents relational errors between adjacent cardiac substructures. Recent progress in whole-heart reconstruction has introduced deep-learning models that generate meshes directly from medical images \cite{kong2021whole,kong2023learningwholeheart,narayanan2023linflonet,meng2023deepmesh,gaggion2023hybridvnet,qiao2025meshheart,beetz2025pointvae}. However, despite these advances, existing methods still lack explicit multi-structure anatomical priors, which are crucial for preserving correct inter-chamber relationships. From an anatomical perspective, the spatial organization of cardiac chambers follows strict inclusion–exclusion principles. The left ventricular (LV) blood pool is fully enclosed by the myocardium (Myo), forming a continuous muscular shell, while the right ventricle (RV) is spatially separated from the LV by the interventricular septum and does not directly intersect with the LV cavity. These hierarchical enclosure and separation relationships constitute fundamental anatomical constraints that must be preserved to ensure physiological plausibility in 3D cardiac reconstruction, as shown in Fig.~\ref{fig1}.

In clinical practice, however, such multi-structure 3D annotations are rarely available. Manual delineation of volumetric cardiac structures is extremely time-consuming and requires highly specialized expertise, making it impractical for large-scale studies \cite{kong2021deep,kong2022learning,wickramasinghe2020voxel2mesh,beetz2022interpretable}. Instead, routine clinical annotation is primarily performed on 2D slices, where individual chambers are carefully delineated and scrutinized in a slice-by-slice manner during diagnosis and quality control \cite{augustin2016anatomically,maher2019accelerating,kong2021whole,kong2023learningwholeheart}. As a result, rich multi-structure information is commonly available in 2D form, whereas fully consistent 3D volumetric labels remain scarce. This modality gap poses a fundamental challenge for enforcing anatomical relationships during 3D mesh reconstruction.

To overcome these limitations, we propose a Mesh Interrelation Enhancement Loss (MIE Loss), a mesh-level objective that explicitly incorporates anatomical priors to enforce spatial consistency. MIE Loss penalizes violations such as surface penetration and unintended inter-chamber connectivity, while leveraging occupancy-based checks to refine mesh deformations and preserve anatomical boundaries. In addition, we introduce two innovative metrics, namely Violation Rate (VR) and Severity-weighted Violation (SVR) as dedicated measurements to quantitatively assess anatomical fidelity. As shown in Fig.~\ref{fig2}, integrating spatial constraints into the reconstruction pipeline produces meshes that are not only geometrically precise but also physiologically coherent, enabling high-fidelity cardiac mesh reconstruction. Our key contributions are summarised as follows:
\begin{itemize}
    \item We propose a novel relational mesh loss (MIE) that explicitly enforces anatomically valid spatial relations between cardiac chambers via a critical-point-guided occupancy formulation.
    \item We introduce a 2D-to-3D relational prior extraction strategy that transforms routine 2D segmentation masks into critical 3D points. These points encode inter-chamber dependencies and enable multi-structure anatomical supervision without requiring volumetric 3D labels.
    \item We develop a unified template-based cardiac mesh reconstruction framework that integrates the proposed MIE loss to jointly regularize surface geometry and inter-chamber anatomical consistency.
    \item We establish two violation-aware relational metrics (VR and SVR) for the direct quantification of inter-chamber anatomical correctness.
\end{itemize}

\section{Relative Work}
 \subsection{Traditional Mesh Deformation in Graphics}

Classical graphics pipelines deform a reference shape by manipulating sparse geometric primitives. Free-form deformation encloses an object within a B-spline lattice whose control points drive global warping  \cite{sederberg1986free}, while cage-based deformation \cite{stroter2024survey} and handle-based techniques \cite{sorkine2004laplacian} adjust vertex displacements through sparse controlling structures. Although effective for animation and shape editing, these approaches require manually designed control structures and do not scale well to anatomically complex multi-object configurations. 

\subsection{Deep Learning Based 3D Shape Reconstruction}
Deep learning has enabled more flexible data-driven mesh reconstruction. Some methods regress 3D surfaces from 2D observations such as RGB images \cite{tang2021skeletonnet}, generative models capture distributional shape priors \cite{liu2021deepmetahandles}, and others transfer deformations between shapes through neural operators \cite{yifan2020neural}. These pipelines typically assume either pre-segmented meshes or coarse 2D inputs, limiting their applicability to complex 3D medical volumes with multiple interacting anatomical structures.

Another dominant line directly predicts vertex-wise offsets on a fixed template. Kong et al. \cite{kong2021deep} regress displacements using GCNs under full 3D supervision, but rely on separate templates per chamber, necessitating post-fusion for whole-heart simulation. Their later work \cite{kong2022learning} uses sparse surface-handle prediction with biharmonic coordinate propagation to generate simulation-ready multi-chamber meshes, but still requires dense volumetric labels for training. Similarly, Voxel2Mesh \cite{wickramasinghe2020voxel2mesh} couples a 3-D CNN encoder with a graph-convolutional mesh decoder that deforms a template directly from volumetric inputs.  Although accurate, these approaches still depend on costly volumetric labels, but control inter-chamber integrity only implicitly through generic smoothness terms, without explicitly encoding multi-structure anatomical priors.

More recent methods further constrain the deformation space. Hui et al. \cite{hui2022neural} enforce homeomorphism to a shared template via per-vertex offsets regularized by edge length and Laplacian smoothness. Beetz et al. \cite{beetz2022interpretable} employ a variational mesh auto-encoder with fixed template connectivity and smoothness priors to maintain spatial consistency. However, the fixed connectivity prevents topology changes, and generic smoothness penalties cannot explicitly enforce chamber-specific anatomical relationships such as LV–RV separation or myocardium enclosure.

Overall, although these template-based methods achieve accurate geometric reconstruction, they rely heavily on dense 3D supervision and regulate inter-chamber consistency only implicitly through generic smoothness or connectivity constraints. As a result, they remain vulnerable to membrane penetration and cross-chamber leakage, particularly in multi-chamber cardiac reconstruction scenarios. Moreover, none of these approaches is explicitly designed to exploit the widely available routine 2D clinical annotations to enforce inter-structure anatomical priors in 3D space.

\begin{figure*}[!t] 
\centering
\centerline{\includegraphics[width=\columnwidth]{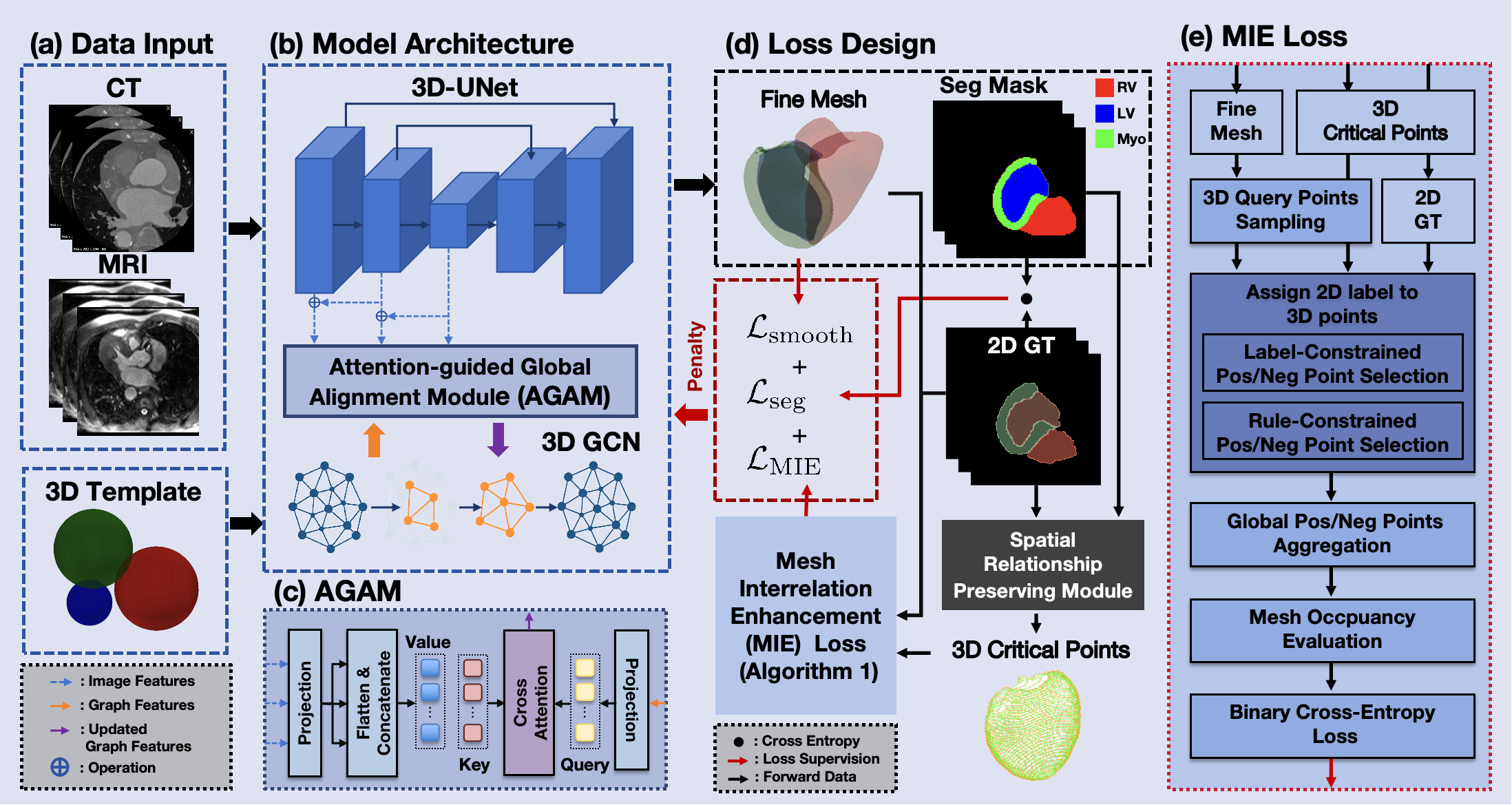}} 
\caption{Overview of the proposed cardiac mesh reconstruction framework with the Mesh Interrelation Enhancement (MIE) loss. 
(a) Multi-modal volumetric images (CT/MRI) and a template as inputs. 
(b) 3D-UNet with an Attention-guided Global Alignment Module (AGAM) and a 3D GCN for mesh deformation. 
(c) Cross-attention based feature alignment between image and mesh features in AGAM. 
(d) Overall training loss integrating smoothness, segmentation, and MIE loss. 
(e) Detailed pipeline of the proposed MIE loss.}
\label{fig2}
\end{figure*}

\section{Methodology}
As illustrated in Fig.~\ref{fig2}, we propose a unified anatomically constrained cardiac mesh reconstruction framework that couples volumetric image representation with template-based mesh deformation under explicit inter-chamber supervision. Given multi-modal CT/MRI volumes and a multi-chamber cardiac template, a 3D-UNet extracts volumetric features, which are aligned with mesh features through an Attention-guided Global Alignment Module (AGAM). The aligned representations are then forwarded to a 3D Graph Convolutional Network (GCN) to predict vertex-wise deformations and generate multi-chamber cardiac meshes.  To enforce anatomically valid inter-chamber relationships, we further introduce the Mesh Interrelation Enhancement (MIE) loss, which derives 3D critical points from routine 2D multi-chamber masks and evaluates relational consistency via a differentiable mesh occupancy formulation. The final objective integrates surface smoothness, segmentation supervision, and the proposed MIE loss, enabling high-fidelity and anatomically consistent cardiac mesh reconstruction without dense 3D volumetric annotations.

\subsection{The architecture of Network}
Following~\cite{kong2021deep}, we first employ a 3D U-Net to extract multi-scale volumetric features for multi-layer segmentation. These features guide a 3D GCN that predicts coarse-to-fine deformations from a template mesh to the target anatomy. Rather than relying on voxel alignment through coordinate interpolation, we introduce AGAM to exchange information between the image features and the graph representation. An initial affine transformation for mesh alignment is regressed from the global deep features at the bottom layer of the U-Net, after which subsequent GCN stages—augmented with image-to-graph attention—predict per-vertex offsets to refine the mesh and enforce anatomical consistency. The network finally outputs a high-fidelity cardiac mesh $\mathcal M$ that is geometrically accurate and anatomically consistent. A detailed illustration of the overall architecture is shown in Fig.~\ref{fig2}.

\subsection{Mesh Inter-Relational Enhancement Loss}
The core idea of the MIE loss is to transform qualitative anatomical relations into quantitative, differentiable supervision signals that act directly on surface geometry. The extracted critical points encode only relational compliance or violation with respect to prescribed spatial rules, rather than absolute 3D surface geometry. Consequently, the proposed loss enforces anatomical consistency through relation-driven supervision, without introducing explicit 3D geometric constraints.
\subsubsection{Spatial Relationship Preserving Module}
Inspired by \cite{gupta2022learning}, the interaction rule set \( \mathcal{R} \) specifies the expected spatial relationships:
\begin{equation}
\mathcal{R} = \{ (A_i, A_j, T_{ij}) \}
\end{equation}
where \(A_i\) and \(A_j\) denote different cardiac structures, and $T_{ij}\in\{0,1\}$. When \(T_{ij}\) = 1, \(A_i\) should be enclosed within \(A_j\), as in the case of the blood pool inside the myocardium. When \(T_{ij}\) = 0, \( A_i\) and \(A_j\) must not overlap, ensuring distinct anatomical boundaries such as the separation of LV and RV, as shown in Fig. \ref{fig1}. For each relationship (\(A_i, A_j\)), the neighborhood overlap of $A_j$ is defined as:
\begin{equation}
N_{A_j} = P(A_j) * K,
\end{equation}
with $P(\cdot)$ the segmentation, $*$ a 3D convolution, and $K$ a 6-connected binary kernel. Violations are then defined as:
\begin{equation}
V_{A_i} = P(A_i)\ \cap\ \bigl((1-T_{ij})\,N_{A_j}\ \cup\ T_{ij}\,(1-N_{A_j})\bigr),
\end{equation}
where \(1 - N_{A_j}\) represents the exterior of \(A_j\). Aggregating over all rules yields the critical voxel map \(V\) which identifies regions that violate the constraints.

\begin{figure}[!t]
\centerline{\includegraphics[width=\columnwidth]{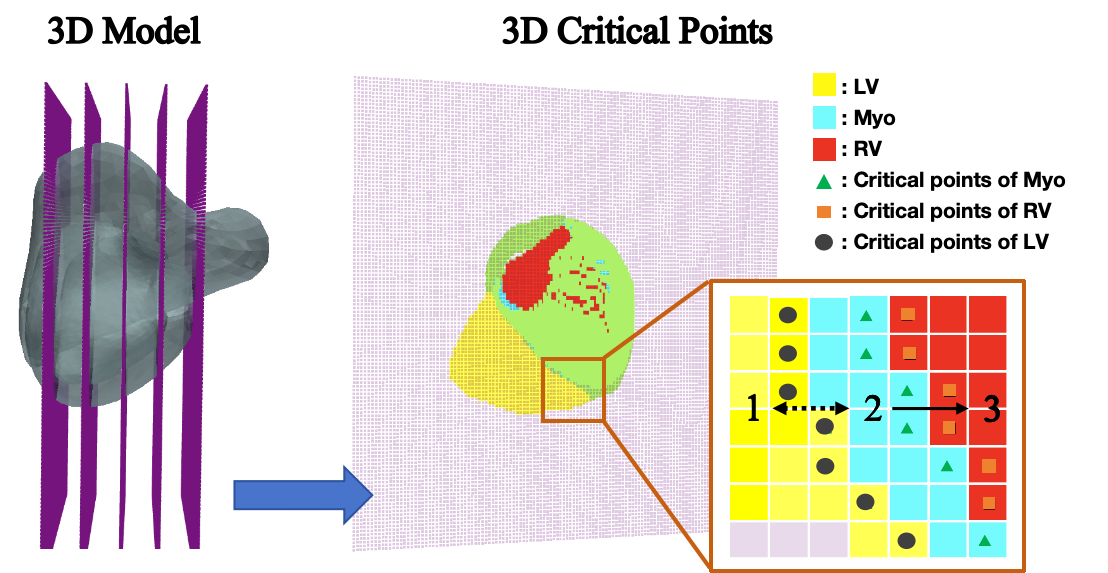}}
\caption{Illustration of the proposed spatial-relationship–preserving module. Anatomical critical points are sampled in the 3D volume to identify regions involved in inter-structure spatial interactions. By enforcing topological consistency over these key regions, the module preserves the prescribed inclusion–exclusion relationships among cardiac structures and effectively suppresses spatial violations during reconstruction. For visualization purposes, only the 3D critical points associated with one representative interaction pair are shown.}
\label{fig3}
\end{figure}

To enable efficient and geometry-aware enforcement of anatomical relational constraints, we convert the voxel-level violation map \(V\) into a point-based representation \(P_{\text{vio}}\) using the image affine transformation \(\mathbf{T}\), as shown in Fig.~\ref{fig3}. While the voxel map encodes violations in a dense grid, a point representation allows us to explicitly localize anatomically critical interfaces in continuous 3D space. This not only avoids redundant volumetric sampling but also enables direct interaction with the surface-based mesh through occupancy evaluation. As a result, the extracted critical points provide a compact and geometrically consistent representation for guiding the subsequent mesh optimization.

\begin{algorithm}[t]
\caption{MIE Loss: Prior knowledge Embedded Mesh Loss }
\label{alg:cpg_loss}
\begin{algorithmic}[1]
\Require
  critical points $P_\text{vio}\!\in\!\mathbb{R}^{M\times3}$\;($x,y,z$),
  mesh $\mathcal M$,
  query points $\mathcal Q\!\in\!\mathbb{R}^{N\times3}$,
  labels $L$,
  interaction rule set $\mathcal{R} = \{ (A_i, A_j, T_{ij}) \}$,
  ratio $\rho$,
  sample size $n$
\Ensure MIE loss $\mathcal L_\text{MIE}$
\State $\mathcal O^{+},\,\mathcal O^{-}\gets[\ ],[\ ]$      \Comment{lists of occupancy scores}
\ForAll{$(A_i, A_j, T_{ij})\in\mathcal R$}           \label{ln:priorBegin}
   \ForAll{$\mathcal M_{A_i} \in \mathcal M $}  \Comment{each class}
      \State $S^+_{A_i}\gets\{q_{xyz}\mid q\in\mathcal Q,\;L[q]=A_i\}$;\;
             $S^-_{A_i}\gets\{q_{xyz}\mid q\in\mathcal Q,\;L[q]\neq A_i\}$
      \If{$T_{ij}=1$}
          \State $S_{p}\!\gets\{p_{xyz}\!\mid\!p\!\in\!P_\text{vio},\;L[p]=A_j\}$
          \State $\mathcal P^{+}= S^+_{A_i}\cup S_p[\Call{Occupancy}{ \mathcal M_{A_i},S_p}=1]$
          \State $\mathcal P^{-}=S^-_{A_i}\cup S_p[\Call{Occupancy}{ \mathcal M_{A_i},S_p}=0]$
      \Else \Comment{Exclusion}
          \State $\mathcal P^{-}=S^-_{A_i}\cup\{p_{xyz}\!\mid\!p\!\in\!P_\text{vio},\;L[point]=A_j\}$
          \State $\mathcal P^{+}= S^+_{A_i}$
      \EndIf \Comment{$\mathcal P^{+},\mathcal P^{-}$contains both routine sampled points and violation-related critical points}
      \State $(o^{+},\,o^{-}) \gets \Call{SplitAndScore}{\mathcal M_{A_i},\;\mathcal P^{+},\;\mathcal P^{-},\;n,\;\rho}$ \Comment{Algorithm 2}
      \State $\Call{Append}{\mathcal O^{+},\,o^{+}};\; \Call{Append}{\mathcal O^{-},\,o^{-}}$
  \EndFor
\EndFor   \label{ln:priorEnd}
\State $\hat{\mathbf y}\!\gets\!\Call{Concat}{\mathcal O^{+},\mathcal O^{-}}$;\quad
       $ \mathbf {GT}\!\gets\![1,\dots,1,0,\dots,0]$
\State $\mathcal L_\text{MIE}\!\gets\!\mathrm{Binary\;Cross\;Entropy}\;(\hat{\mathbf y},\mathbf {GT})$
\State \Return $\mathcal L_\text{MIE}$
\end{algorithmic}
\end{algorithm}

\subsubsection{Critical Interaction Mesh Optimization} Once the critical points $P_{\text{vio}}$ have been localized, we use them to drive the mesh optimization via a critical-point-guided occupancy formulation, as shown in Algorithm~\ref{alg:cpg_loss}. We assign each critical point a binary target according to the prescribed spatial relationship: points that satisfy the prescribed relation are labelled as compliant, whereas violating points receive a penalising label. We build upon the differentiable occupancy formulation \(O(p)\), which is used to score both uniformly sampled queries and the detected critical points. This extends the standard occupancy formulation by embedding chamber-wise inclusion–exclusion priors, which helps enforce anatomically consistent mesh deformation.

The key idea of the differentiable occupancy function \(O(p)\) \cite{luo2024explicit} is to estimate whether a set of query points \(p \in \mathcal{Q}\) is contained inside of a cardiac structure by,
\begin{equation}
O(p) = \frac{1}{4\pi} \sum_{v \in V} A_v \cdot \frac{(p - v)}{\| p - v \|^3},
\end{equation}
where \(A_v\) is the dual area weight of vertex \(v\), \(\| p - v \|^3\) represents the cubed Euclidean distance between \(p\) and \(v\), the factor $1/(4\pi)$ ensuring that the result approximates $0$ or $1$.

For a target structure $A_i$, query points \(p \in \mathcal{Q}\) are randomly sampled from/near mesh and classified as positive or negative. For example, negative samples not only include points from other structures or the background but also additional spatially-aware modifications to enforce anatomical consistency. 
\begin{equation} 
S^+_{A_i} = \{ p \in \mathbb{R}^3 \mid L(p) = A_i \}, S^-_{A_i} = \{ p \in \mathbb{R}^3 \mid L(p) \neq A_i \},
\end{equation}
where \(L(p)\) represents the ground truth label at point \(p\), \(S^+_{A_i}\) denotes the positive sample set, and \(S^-_{A_i}\) represents the negative sample set.
\paragraph{Exclusion Constraints}
An exclusion constraint ensures that \(A_i\) and \(A_j\) remain spatially disjoint, preventing overlap. Violations are added to the negative sample set, while positive samples remain unchanged, consisting of standard points from \(A_i\). The negative sample set includes both global negatives and overlapping points from \(A_j\).
\paragraph{Inclusion Constraints}
An inclusion constraint ensures that \(A_i\) is fully contained within \(A_j\), preventing boundary violations. Sample selection is adjusted accordingly, with positive samples including both standard points from \(A_i\) and enclosing points from \(A_j\).
Negative samples include points from outside \(A_i\) and additional points from \(A_i\) that incorrectly extend beyond \(A_j\).

All rule-constraints positive and negative point sets are then processed by the mixed sampling and occupancy scoring procedure (Algorithm~\ref{alg:sample_score}), and the resulting occupancy scores are aggregated to form the final MIE loss via binary cross-entropy, as specified in Algorithm~\ref{alg:cpg_loss}. The choice of the sampling size $n$ and the critical sampling ratio $\rho$ is discussed in detail in Section~\ref{sec:efficiency}.

\subsubsection{Loss for Mesh Optimization}
The MIE Loss optimizes the reconstructed mesh by enforcing anatomical validity through constraints and binary cross-entropy (BCE) loss. We adopt the BCE loss because the occupancy formulation naturally defines a per-point binary classification problem, for which BCE provides a direct and well-calibrated optimization objective. Compared with region-based losses, such as the Dice loss, which measure set-level overlap, BCE operates at the level of individual query points and is therefore more sensitive to sparse but anatomically critical violations, such as local penetration and thin leakage. In addition, BCE is known to provide stable gradients under severe class imbalance, which is common in relational sampling, where violation points constitute only a small fraction of all query points. These properties make BCE particularly well-suited for robust optimization of point-wise relational mesh constraints.
\begin{equation}
\mathcal{L}_{\text{MIE}} = - \frac{1}{N} \sum_{p} O_{\text{gt}}(p) \log O(p) + (1 - O_{\text{gt}}(p)) \log (1 - O(p)),
\end{equation}
where \(O(p)\) is the predicted occupancy, \(O_{\text{gt}}(p)\) is the ground truth. The ground truth occupancy follows:
\begin{equation}
O_{\text{gt}}(p) =
\begin{cases}
1, & p \in \mathcal P^+ \\
0, & p \in \mathcal P^-.
\end{cases}
\end{equation}
\subsubsection{Computational Efficiency}
\label{sec:efficiency}

Evaluating the loss on every candidate point is prohibitively expensive on the GPU. Therefore, we subsample a fixed budget of $n=2000$ query points per iteration, as described in Algorithm~\ref{alg:sample_score}. To ensure that a portion of these samples consistently falls on anatomically sensitive interfaces, each positive and negative set $\mathcal P^{+}$ and $\mathcal P^{-}$ is first split into a violation-critical subset $P_{\text{vio}}^{+}, P_{\text{vio}}^{-}$ and a regular subset $S^{+}, S^{-}$, respectively. We then uniformly sample $\rho n\;(=0.2n)$ points from $P_{\text{vio}}^{+}$ and $P_{\text{vio}}^{-}$, and the remaining $(1-\rho)n$ points from $S^{+}$ and $S^{-}$. All selections are performed uniformly at random without replacement. The choice $\rho=0.2$ reflects a practical compromise that keeps the memory footprint within the 2000-point budget while ensuring that each mini-batch still contains a sufficient number of interface samples to effectively detect and penalize anatomical prior violations. As shown later in Section~\ref{sec:experiments}, allocating the entire sampling budget to regular points ($\rho=0$) leads to a noticeable reduction in prior consistency.

\subsection{Overall Loss}

We formulate a unified loss function to supervise mesh deformation under different experimental settings:
\begin{equation}
\mathcal{L} = 
\lambda_{\text{chamfer}} \mathcal{L}_{\text{chamfer}}+
\lambda_{\text{Occ}} \mathcal{L}_{\text{Occ}} 
+ \mathcal{L}_{\text{seg}} 
+ \lambda_{\text{MIE}} \mathcal{L}_{\text{MIE}} 
+  \mathcal{L}_{\text{smooth}} .
\end{equation}

In the primary training setting illustrated in Fig.~\ref{fig2}, dense 3D mesh ground truth and explicit voxel-wise occupancy annotations are not available. Therefore, we set $\lambda_{\text{chamfer}} = 0$ and $\lambda_{\text{Occ}} = 0$, and optimize the network using:
\begin{equation}
\mathcal{L}_{\text{train}} = 
\mathcal{L}_{\text{seg}} 
+ \mathcal{L}_{\text{MIE}} 
+ \mathcal{L}_{\text{smooth}} .
\end{equation}

\paragraph{Segmentation Loss}
$\mathcal{L}_{\text{seg}}$ provides pixel-wise 2D multi-class segmentation supervision.
\paragraph{Chamfer Distance}
When 3D ground-truth meshes are available, the Chamfer distance supervises geometric accuracy between the predicted mesh $\mathcal{M}$ and the reference mesh \(\Omega_G\).
\paragraph{Occupancy-based Loss}
The occupancy loss $\mathcal{L}_{\text{Occ}}$ and the proposed MIE loss operate on point-wise 3D regular points $p \in \mathcal{Q}$ and adopt a unified binary cross-entropy formulation:
\begin{equation}
\begin{aligned}
\mathcal{L}_{\text{Occ}}(\mathcal{Q}) =
- \sum_{p \in \mathcal{Q}}
\Big[
& O_{\text{gt}}(p)\log O(p) \\
& + (1-O_{\text{gt}}(p))\log(1-O(p)) 
\Big].
\end{aligned}
\end{equation}

\begin{algorithm}[H]
\caption{\textsc{SplitAndScore}$(\mathcal M,\;\mathcal P^{+},\;\mathcal P^{-},\;n,\;\rho=0)$}
\label{alg:sample_score}
\begin{algorithmic}[1]
\Require mesh $\mathcal M$; positive / negative point sets $\mathcal P^{+},\mathcal P^{-}$;
        sample size $n$; ratio $\rho\!\in[0,1]$ (default $0$)
\Ensure $(o^{+},o^{-})$ — occupancy scores for positive / negative samples
\State $(P_\text{vio}^{+},\,S^{+}) \gets \Call{Split}{\mathcal P^{+}}$
\State $(P_\text{vio}^{-},\,S^{-}) \gets \Call{Split}{\mathcal P^{-}}$
\Comment Split each set into violation-critical points and regular points

\If{$\rho = 0$}
    \State $R^{+}\gets\Call{PickRandom}{\mathcal P^{+},\,n}$
    \State $R^{-}\gets\Call{PickRandom}{\mathcal P^{-},\,n}$
\Else
    \State $m\gets\min\!\bigl(|P_\text{vio}^{+}|,\;|P_\text{vio}^{-}|,\;\lfloor\;\rho n\;\rfloor\bigr)$
    \Comment number of critical points sampled from each set
    \State $R^{+}\gets\Call{PickRandom}{P_\text{vio}^{+},m}\ \cup\
           \Call{PickRandom}{S^{+},\,n-m}$
    \State $R^{-}\gets\Call{PickRandom}{P_\text{vio}^{-},m}\ \cup\
           \Call{PickRandom}{S^{-},\,n-m}$
\EndIf

\State $o^{+}\gets Sigmoid\bigl(\Call{Occupancy}{\mathcal M,R^{+}}\bigr)$
\State $o^{-}\gets Sigmoid\bigl(\Call{Occupancy}{\mathcal M,R^{-}}\bigr)$
\State \Return $(o^{+},o^{-})$
\end{algorithmic}
\end{algorithm}

\paragraph{Surface Smoothness}
The smoothness loss regularizes the local surface geometry using Laplacian coordinates:
\begin{equation}
\mathcal{L}_{\text{smooth}} = 
\frac{1}{|\mathcal{V}|} 
\sum_{v \in \mathcal{V}} 
\left\| 
v - \frac{1}{|\mathcal{N}(v)|}
\sum_{u \in \mathcal{N}(v)} u 
\right\|^2 .
\end{equation}
Here, $\mathcal{V}$ denotes the set of all mesh vertices, $v \in \mathcal{V}$ is an individual vertex, and $\mathcal{N}(v)$ represents the one-ring neighborhood of $v$, i.e., the set of vertices directly connected to $v$ by mesh edges, with $|\mathcal{V}|$ and $|\mathcal{N}(v)|$ denoting their respective cardinalities. This term penalizes deviations of each vertex from the centroid of its one-ring neighborhood, thereby encouraging locally smooth, non-noisy, and anatomically plausible surface deformations while preventing irregular or spiky artifacts during optimization.

\section{Experimental analysis}
\label{sec:experiments}
For the baseline, we extract the reference surface mesh \(\Omega_G\) by marching Cubes algorithm \cite{cline19873d} from the ground truth segmentation first, and use chamfer distance from reconstructed mesh to reference as the main supervision. For the ablation study, we train the reconstruction network by the   
differentiable rendering \cite{luo2024differentiable} with random sampling and edge-enhanced sampling but no MIE loss involved, noted as $\mathcal L_\text{Occ}$ and $\mathcal L^{0.2}_\text{Occ}$ where $0.2$ means 20\% extra sample from edge area, following the sampling strategy described in Algorithm~\ref{alg:sample_score}. Our experiments show that even edge-enhanced sampling cannot address the intersection problem in multi-chambers, underscoring the importance of the newly proposed $MIE$.


\begin{table*}[t]
\centering
\caption{Summary of data characteristics for CT and MR datasets.}
\label{tab:data_summary}
\resizebox{\linewidth}{!}{
\begin{tabular}{lcccccc}
\toprule
\multicolumn{1}{c}{\textbf{Dataset}} &
\multicolumn{3}{c}{\textbf{CT data}} &
\multicolumn{3}{c}{\textbf{MR data}}\\
\cmidrule(lr){2-4}\cmidrule(l){5-7}
 & MMWHS\cite{zhuang2019evaluation} & WHS++\cite{zhang2024preserving} & ImageCAS\cite{zeng2023imagecas} & MMWHS\cite{zhuang2019evaluation} & WHS++\cite{zhang2024preserving}& HVSMR\cite{pace2024hvsmr} \\
\midrule
Vendor & Philips & GE/Philips/Siemens & Siemens  & 1.5T Philips \& Siemens & 1.5T Philips \& Siemens  & 1.5T Philips \\
\# clinical sites & 2 & 6 & 1 & 1 & 2 & 1  \\
\# 3-D volumes & 60 & 104 & 1000  & 60 & 102 & 26 \\
\# patients & 60 & 72 & 1000  & 60 & 27 & 26 \\
In-plane res. (mm) & 0.78 $\times$ 0.78 & 0.40–0.50 $\times$ 0.40–0.50 & 0.25 $\times$ 0.25  & 1.60–2.00 $\times$ 1.60–2.00 & 1.25 $\times$ 1.25 & 0.73 $\times$ 0.73 \\
Slice thickness (mm) & 1.6 & 0.5–0.625 & 0.25-0.45  & 2.0–3.2 & 2.7 & 0.81 \\
\bottomrule
\end{tabular}}
\end{table*}
\subsection{Datasets}
We evaluate MIE Loss on multi-modal datasets (CT: MMWHS \cite{zhuang2019evaluation}, WHS++ \cite{zhang2024preserving}, ImageCAS \cite{zeng2023imagecas}; MRI: MMWHS \cite{zhuang2019evaluation}, WHS++ \cite{zhang2024preserving}) with varying resolutions, as shown in Table~\ref{tab:data_summary}. CT datasets provide higher-resolution structural details, while MRI datasets exhibit greater anatomical variability, making them ideal for assessing generalization. Training was performed separately on multi-center CT and MR datasets from MMWHS and WHS++, ensuring modality-specific learning.

\begin{figure*}[t]
    \centering
        \subfigure[Top-10 congenital pathologies]{
        \includegraphics[width=0.28\textwidth]{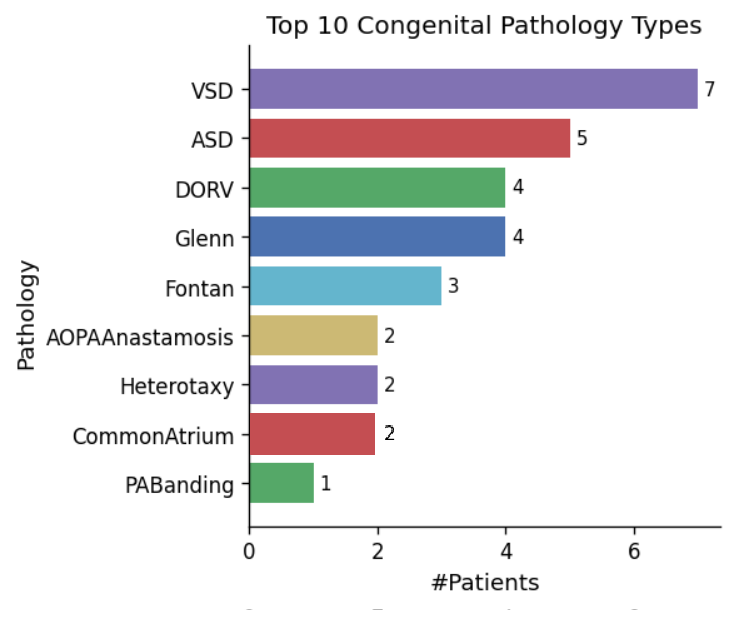}
        \label{fig:pathology_top10}
    }  
    \hfill
    \subfigure[Age distribution]{
        \includegraphics[width=0.32\textwidth]{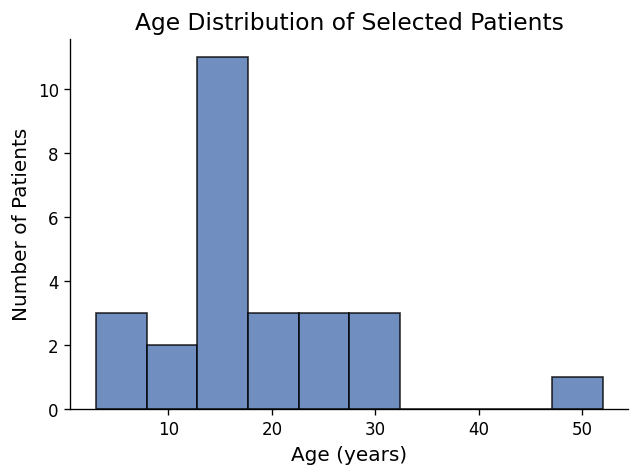}
        \label{fig:age_hist}
    }
     \hfill
    \subfigure[Severity distribution]{
        \includegraphics[width=0.25\textwidth]{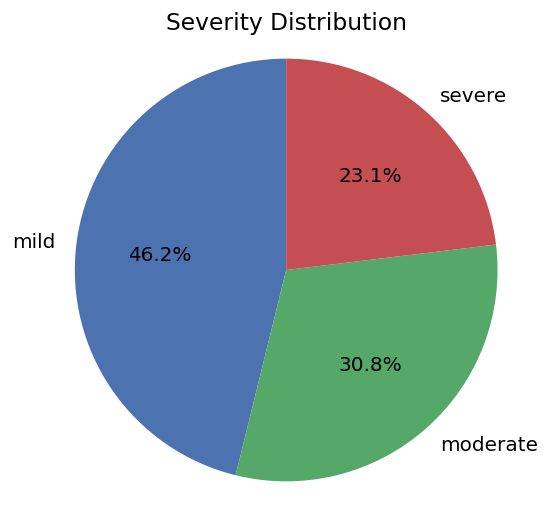}
        \label{fig:severity_pie}
    }    
    \caption{
        Clinical heterogeneity of the selected HVSMR congenital heart disease cohort.
        (a) Prevalence of the top-10 congenital structural abnormalities, highlighting substantial phenotypic diversity. 
        (b) Age histogram revealing the wide age span (0--52 years), indicating varied anatomical scales.  
        (c) Severity-level distribution (mild/moderate/severe) among the 26 subjects included in our study.
    }
    \label{fig:mit_clinical_distribution}
\end{figure*}

External inference was further conducted on two independent cohorts—the ImageCAS CT dataset \cite{zeng2023imagecas} and the HVSMR congenital heart disease MRI cohort \cite{pace2024hvsmr}—to evaluate robustness across imaging centers and modalities, as well as the model’s behavior under substantial anatomical variability characteristic of congenital heart disease (CHD). The original HVSMR collection contains a diverse set of CHD MRI scans; however, only 26 subjects were included in our reconstruction study due to data completeness and anatomical consistency constraints. Specifically, these cases were selected as those for which both the raw MR volumes and reliable, chamber-consistent manual annotations of the left and right ventricles (LV and RV) were simultaneously available. The remaining scans either exhibited incomplete label coverage, ambiguous ventricular identity (e.g., single-ventricle physiology), or poor image quality that precluded their use in multi-chamber reconstruction. Despite the reduced sample size, the retained cohort remains highly heterogeneous, spanning a broad spectrum of congenital abnormalities, age distributions, and severity grades (Fig.~\ref{fig:mit_clinical_distribution}). This variability ensures that the evaluation reflects realistic bi-ventricular CHD phenotypes rather than a homogenized subset.

To obtain an objective assessment, we manually segmented the original image-only 60 CT + 60 MR test scans on the WHS++ data: two radiologists with over 5 years of experience in cardiovascular image labelling independently performed the initial labelling, which was then uniformly reviewed by a third senior physician to generate the final true value masks in accordance with the official structural labels (LV, RV, Myo). These manual labels are only used for final performance reporting and are not involved in any training or tuning process. The surface quality was evaluated in terms of surface smoothness, normal consistency and spatial correctness.

\subsection{Evaluation Metric}
We assess reconstruction quality using the Dice Similarity Coefficient (DSC), Chamfer Distance (CD), and Hausdorff Distance (HD) between the 3D ground truth and the predicted mesh (Eqs.~\ref{eq:dice}--\ref{eq:hd}). DSC measures segmentation accuracy via volumetric overlap with the ground truth, CD quantifies average geometric deviation using nearest-point distances, and HD captures worst-case surface discrepancy. We also report a Laplacian smoothness energy (LSE) to evaluate mesh regularity (Eq.~\ref{eq:lap}). 

Unlike these geometry-based metrics, we design  two complementary indicators. VR measures the fraction of critical points that contradict a given spatial relation, taking values in [0,1] in which 0 indicates full compliance and 1 indicates that essentially all candidates violate the relation. SVR measures how severe those violations are by averaging the signed margin to the decision threshold over the violating points; larger SVR means stronger inconsistency—farther outside for inclusion relations and deeper intrusion for exclusion relations(Eq.~\ref{eq:I}-~\ref{eq:svr}). 
\paragraph{VR and SVR} Let $\mathrm{Mesh}_i$ be the predicted mesh for class $i$, $P_{\mathrm{vio}}^{\,j}$ denotes the critical-point set associated with class $j$ for relation $(i,j,T_{ij})$.
\begin{equation}
     I_{ij} = \sigma( \mathrm{Occupancy}(\mathcal{M}_{i}, P_{\text{vio}}^j ) ),
\label{eq:I}
\end{equation}
\begin{equation}
\mathrm{VR}\;=\displaystyle\sum_{(i,j,T_{ij})\in\mathcal R}\frac{ 
|(1-T_{ij})\,I_{ij}\;+\;T_{ij}\,\big(1-I_{ij}\big)|}
{\displaystyle\ |P_{\text{vio}}^j|},
\label{eq:vr}
\end{equation}
\begin{equation}
\mathrm{SVR}\;=\displaystyle\sum_{(i,j,T_{ij})\in\mathcal R} 
(1-T_{ij})\,I_{ij}\;+\;T_{ij}\,\big(1-I_{ij}\big),
\label{eq:svr}
\end{equation}
where $|\cdot|$ its cardinality; $\sigma(\cdot)$ is the sigmoid.
\paragraph{Dice Similarity Coefficient (DSC)} Given volumetric occupancies \(\Omega_M\), the DSC is
\begin{equation}
\mathrm{DSC}(\Omega_M,\Omega_G)=
\frac{2\,|\Omega_M \cap \Omega_G|}{|\Omega_M|+|\Omega_G|}.
\label{eq:dice}
\end{equation}
\paragraph{Chamfer Distance (CD)} Let \(P=\{p_k\}_{k=1}^{|P|}\) and \(Q=\{q_\ell\}_{\ell=1}^{|Q|}\) be points sampled from the predicted and reference surfaces, respectively. The (squared) symmetric Chamfer distance is
\begin{equation}
\mathrm{CD}(P,Q)=
\frac{1}{|P|}\sum_{p\in P}\min_{q\in Q}\|p-q\|_2^{2}
\;+\;
\frac{1}{|Q|}\sum_{q\in Q}\min_{p\in P}\|q-p\|_2^{2}.
\label{eq:cd}
\end{equation}

\paragraph{Hausdorff Distance (HD)} The symmetric Hausdorff distance between the same point sets \(P,Q\) is
\begin{equation}
\mathrm{HD}(P,Q)=
\max\!\left\{
\;\max_{p\in P}\min_{q\in Q}\|p-q\|_2,\;
\max_{q\in Q}\min_{p\in P}\|q-p\|_2
\right\}.
\label{eq:hd}
\end{equation}

\paragraph{Laplacian Smoothness Energy (LSE)} For the predicted mesh \(\mathcal{M}=(V,E)\), define the (cotangent) Laplacian at vertex \(v_i\) as
\[
\Delta v_i
= v_i - \frac{\sum_{j\in\mathcal{N}(i)} w_{ij}\, v_j}{\sum_{j\in\mathcal{N}(i)} w_{ij}},
\quad
w_{ij}=\tfrac{1}{2}\!\left(\cot\alpha_{ij}+\cot\beta_{ij}\right),
\]
where \(\alpha_{ij},\beta_{ij}\) are the angles opposite edge \((i,j)\) in the two incident triangles. The (mean) Laplacian smoothness energy is
\begin{equation}
E_{\mathrm{Lap}}(\mathcal{M})=
\frac{1}{|V|}\sum_{i\in V}\|\Delta v_i\|_2^{2},
\label{eq:lap}
\end{equation}
with smaller values indicating a smoother surface.

\begin{table*}[t]
\centering
\scriptsize
\setlength{\tabcolsep}{4pt}
\caption{Performance on the WHS++ CT and MR test sets.
         Higher\,=\,better for Dice/Jaccard; lower\,=\,better for ASSD/HD/VPS.
$^{*}$ p < 0.05 vs $\mathcal L_{\text{Chamfer}}$;\quad
$^{\dagger}$ p < 0.05 vs $\mathcal L_{\text{Occ}}$.
}
\label{tab:whs_full}
\begin{tabular}{lllcccccc}
\toprule
\textbf{Mod.} & \textbf{Struct.} & \textbf{Method} &
\textbf{DSC} $\uparrow$ & \textbf{CD (mm)} $\downarrow$ &
\textbf{HD (mm)} $\downarrow$ & \textbf{LSE} $\downarrow$ &
\textbf{VR} $\downarrow$ & \textbf{SVR} $\downarrow$\\
\midrule
\multirow{20}{*}{CT}
 & \multirow{5}{*}{LV}
   & $\mathcal L_{\text{Chamfer}}$               & \textbf{0.9227±0.0825} & 0.0053±0.0029 & 0.1663±0.0396 & \textbf{0.2902±0.0140} & 0.1998±0.0104  & 0.1968±0.0020\\
 & & $\mathcal L_{\text{Occ}}$                   &  0.8865±0.0938$^{*}$   & 0.0017±0.0003$^{*}$ & 0.0889±0.0324$^{*}$  & 0.3126±0.0271$^{*}$  & 0.1974±0.0034 & 0.1979±0.0014\\
 & & $\mathcal L_{\text{Occ}}^{0.2}$     & 0.5045±0.1314    & 0.0271±0.0145    & 0.1876±0.0451    & 0.2996±0.0273    & 0.4902±0.2258 & 0.4636±0.0258\\
 & & $\mathcal L_{\text{MIE}}$                   & 0.9198±0.0886$^{\dagger}$ & \textbf{0.0012±0.0002}$^{*}$$^{\dagger}$ & \textbf{0.0798±0.0364}$^{*}$$^{\dagger}$ & 0.3021±0.0268$^{\dagger}$ & \textbf{0.0926±0.0026}$^{*}$$^{\dagger}$ & 0.1056±0.0022$^{*}$$^{\dagger}$\\
 & & $\mathcal L_{\text{MIE}}^{0.2}$     & 0.9051±0.0749    & 0.0015±0.0001    & 0.0879±0.0322    & 0.2999±0.0267   & 0.0984±0.0022 & \textbf{0.0957±0.0024} \\
\cmidrule(lr){2-9}
 & \multirow{5}{*}{RV}
   & $\mathcal L_{\text{Chamfer}}$               & \textbf{0.9064±0.06226} & 0.0052± 0.0028& 0.1598±0.0494 & 0.2651±0.0165 & 0.0923±0.0368 & 0.2288±0.0046 \\
 & & $\mathcal L_{\text{Occ}}$                   & 0.8403±0.0730$^{*}$    & 0.0021±0.0015$^{*}$ & 0.1163±0.0339$^{*}$    & 0.2625±0.0274    & 0.1177±0.0296 & 0.2265±0.0036\\
 & & $\mathcal L_{\text{Occ}}^{0.2}$     & 0.7474±0.0667   & \textbf{0.0018±0.0012}    & \textbf{0.1119±0.0344}    & 0.2877±0.0294    & 0.2444±0.0632 & 0.2276±0.0048\\
 & & $\mathcal L_{\text{MIE}}$                   & 0.8498±0.0641$^{*}$$^{\dagger}$ & 0.0019±0.0009$^{*}$ & 0.1166±0.0286$^{*}$ & 0.2641±0.0271$^{\dagger}$ & \textbf{0.0153±0.0156}$^{*}$$^{\dagger}$  & \textbf{0.0939±0.0213}$^{*}$$^{\dagger}$ \\
 & & $\mathcal L_{\text{MIE}}^{0.2}$     & 0.8405±0.0648    & 0.0019±0.0011    & 0.1155±0.0282    & \textbf{0.2621±0.0269}    & 0.0234±0.0186 & 0.0947±0.0183\\
\cmidrule(lr){2-9}
 & \multirow{5}{*}{Myo}
& $\mathcal L_{\text{Chamfer}}$               & \textbf{0.8996±0.1361} & 0.0099±0.0036 & 0.1852±0.0363 & 0.2418±0.0185  & 0.2512±0.0343& 0.2273±0.0023 \\
 & & $\mathcal L_{\text{Occ}}$                   & 0.8086±0.140$^{*}$& 0.0013±0.0014$^{*}$ & 0.1074±0.0411$^{*}$ & 0.2339±0.0194$^{*}$ & 0.2459±0.0251 & 0.2248±0.0028\\
 & & $\mathcal L_{\text{Occ}}^{0.2}$     & - & \textbf{ 0.0006±0.0003} &\textbf{ 0.0695±0.0254} & 0.2579±0.0211 & 0.41494±0.0884 & 0.2449±0.0223\\
 & & $\mathcal L_{\text{MIE}}$                   & 0.8444±0.1418$^{*}$$^{\dagger}$ & 0.0014±0.0014$^{*}$ & 0.1001±0.0435$^{*}$ & 0.2349±0.0198$^{*}$ & 0.1074±0.0084$^{*}$$^{\dagger}$  & 0.1978±0.0022$^{*}$$^{\dagger}$ \\
 & & $\mathcal L_{\text{MIE}}^{0.2}$     & 0.8265±0.0749 & 0.0014±0.0015 & 0.1067±0.0425 & \textbf{0.2324±0.0195} & \textbf{0.0975±0.0051} & \textbf{0.1221±0.0024}\\
\cmidrule(lr){2-9}
 & \multirow{5}{*}{Sum}
& $\mathcal L_{\text{Chamfer}}$               & \textbf{0.9095±0.0860} & 0.0204± 0.0046 & 0.5114±0.0755 & 0.8071±0.0477 & 0.1544±0.0090& 0.2018±0.0025 \\
 & & $\mathcal L_{\text{Occ}}$                   & 0.8454±0.0940$^{*}$ & 0.0050±0.0039$^{*}$ & 0.3126±0.0758$^{*}$  & 0.8090±0.0735 & 0.2041±0.0301$^{*}$& 0.2099±0.0021 \\
 & & $\mathcal L_{\text{Occ}}^{0.2}$     & 0.4676±0.0614 & 0.0296±0.0147 & 0.3691±0.0771 & 0.8452±0.0775 & 0.2815±0.1446 & 0.2823±0.0094\\
 & & $\mathcal L_{\text{MIE}}$                   & 0.8617±0.0898$^{*}$$^{\dagger}$ & \textbf{0.0045±0.0035}$^{*}$ & \textbf{0.3097±0.0824}$^{*}$ & 0.8011±0.0732$^{*}$$^{\dagger}$ & 0.0545±0.0086$^{*}$$^{\dagger}$  & 0.1675±0.0898$^{*}$$^{\dagger}$ \\
 & & $\mathcal L_{\text{MIE}}^{0.2}$     & 0.8604±0.0745 & 0.0047±0.0034 & 0.3150±0.0696 & \textbf{0.7944±0.0727} & \textbf{0.0478±0.0067} & \textbf{0.1179±0.0077}\\
\midrule
\multirow{20}{*}{MR}
 & \multirow{5}{*}{LV}
   & $\mathcal L_{\text{Chamfer}}$               & 0.8012±0.0765 & 0.0039±0.0021 & 0.1441±0.0316 & \textbf{0.2652±0.0183} & 0.1909±0.0204 & 0.4979±0.0011\\
 & & $\mathcal L_{\text{Occ}}$                   & 0.8609±0.0952$^{*}$ & 0.0038±0.0026 & 0.1404±0.0338 & 0.3281±0.0208$^{*}$ & 0.1949±0.0067 & 0.4976±0.0015 \\
 & & $\mathcal L_{\text{Occ}}^{0.2}$     & 0.5037±0.0967 & 0.0163±0.0044 & 0.2717±0.0378 & 0.3446±0.0265 & 0.6014±0.1577 & 0.4487±0.0292 \\
 & & $\mathcal L_{\text{MIE}}$                   & \textbf{0.8657±0.0988}$^{*}$ & \textbf{0.0029±0.0028}$^{*}$$^{\dagger}$ & \textbf{0.1291±0.0339}$^{*}$$^{\dagger}$ & 0.3366±0.0298$^{*}$ & 0.1256±0.0041$^{*}$$^{\dagger}$  & 0.2552±0.0020 $^{*}$$^{\dagger}$ \\
 & & $\mathcal L_{\text{MIE}}^{0.2}$     & 0.8472±0.1034 & 0.0036±0.0032 & 0.1525±0.0375 & 0.3363±0.0288 & \textbf{0.0978±0.0026} & \textbf{0.1961±0.0018}\\
\cmidrule(lr){2-9}
 & \multirow{5}{*}{RV}
   & $\mathcal L_{\text{Chamfer}}$               & 0.6672±0.1141 & 0.0076±0.0039 & 0.2204±0.0516 & \textbf{0.2325±0.0186} & 0.0617±0.0969 & 0.2053±0.0703\\
 & & $\mathcal L_{\text{Occ}}$                   & 0.7399±0.1750$^{*}$ & 0.0082±0.0111$^{*}$ & 0.1947±0.0907$^{*}$ & 0.2955±0.0324$^{*}$ & 0.1071±0.0271$^{*}$ & 0.2192±0.0056\\
 & & $\mathcal L_{\text{Occ}}^{0.2}$     & 0.7143±0.1625 & 0.0088±0.0110 & 0.2021±0.0914 & 0.3160±0.0273 & 0.2029±0.0518  & 0.2332±0.0053 \\
 & & $\mathcal L_{\text{MIE}}$                   & \textbf{0.7669±0.1651}$^{*}$$^{\dagger}$ & \textbf{0.0077±0.0103}$^{\dagger}$ & \textbf{0.1747±0.0813}$^{*}$$^{\dagger}$ & 0.3047±0.0327$^{*}$  & 0.0129±0.0141 $^{*}$$^{\dagger}$ & 0.1985±0.0209$^{*}$$^{\dagger}$ \\
 & & $\mathcal L_{\text{MIE}}^{0.2}$     & 0.7490±0.1605 & 0.0079±0.0094 & 0.1887±0.0906 & 0.3026±0.0322 & \textbf{0.0117±0.0067} & \textbf{0.1203±0.0181}\\
\cmidrule(lr){2-9}
 & \multirow{5}{*}{Myo}
   & $\mathcal L_{\text{Chamfer}}$               & 0.7071±0.1209 & 0.0060±0.0041 & 0.1911±0.0585 & \textbf{0.2164±0.0146} & 0.1470±0.0354 & 0.2286±0.0030\\
 & & $\mathcal L_{\text{Occ}}$                   & 0.7403±0.1218$^{*}$ & 0.0019±0.0010$^{*}$ & 0.1433±0.0319$^{*}$ & 0.2599±0.0196$^{*}$ &0.1567±0.0187$^{*}$ & 0.2216±0.0041\\
 & & $\mathcal L_{\text{Occ}}^{0.2}$     & 0.3651±0.1499 & 0.0029±0.0016 & 0.1860±0.0451 & 0.2727±0.0175 & 0.1394±0.0536 & 0.2505±0.0196\\
 & & $\mathcal L_{\text{MIE}}$                   & \textbf{0.7414±0.1319}$^{*}$ & \textbf{0.0014±0.0013}$^{*}$$^{\dagger}$ & \textbf{0.1405±0.0335}$^{*}$ & 0.2672±0.0215$^{*}$ & 0.1145±0.0104 $^{*}$$^{\dagger}$ & \textbf{0.1895±0.0040}$^{*}$$^{\dagger}$  \\
 & & $\mathcal L_{\text{MIE}}^{0.2}$     & 0.7222±0.1340 & 0.0019±0.0011 & 0.1579±0.0386 & 0.2651±0.0195 & \textbf{0.0962±0.0044} & 0.1923±0.0041\\
\cmidrule(lr){2-9}
 & \multirow{5}{*}{Sum}
   & $\mathcal L_{\text{Chamfer}}$               & 0.6115±0.0595 & 0.0175±0.0061 & 0.5558±0.0819 & \textbf{0.7143±0.0509} & 0.2397±0.0850 & 0.2726±0.0275\\
 & & $\mathcal L_{\text{Occ}}$                   & 0.7688±0.1184$^{*}$ & 0.0139±0.0139$^{*}$ & 0.4784±0.1364$^{*}$ & 0.8833±0.0795$^{*}$ & 0.2787±0.0580$^{*}$ & 0.2536±0.0032\\
 & & $\mathcal L_{\text{Occ}}^{0.2}$     & 0.6414±0.1034 & 0.0281±0.0153 & 0.6598±0.1376 & 0.9333±0.0703 & 0.3754±0.0863 & 0.2805±0.0054\\
 & & $\mathcal L_{\text{MIE}}$                   & \textbf{0.7904±0.1196}$^{*}$$^{\dagger}$ & \textbf{0.0120±0.0133}$^{*}$$^{\dagger}$ & \textbf{0.4443±0.1192}$^{*}$$^{\dagger}$ & 0.9086±0.0833$^{*}$$^{\dagger}$ & 0.0515±0.0625 $^{*}$$^{\dagger}$  & 0.1876±0.0091$^{*}$$^{\dagger}$ \\
 & & $\mathcal L_{\text{MIE}}^{0.2}$     & 0.7852±0.1187 & 0.0134±0.0125 & 0.4584±0.1507 & 0.9039±0.0799 & \textbf{0.0423±0.0524 }& \textbf{0.1443±0.0082}\\
\bottomrule

\end{tabular}
\end{table*}

\subsection{Results}
All statistical comparisons were conducted using paired two-sided Wilcoxon signed-rank tests at the case level. Differences were considered statistically significant when $p < 0.05$. In the result tables, $^{*}$ and $^{\dagger}$ denote statistically significant differences compared with $\mathcal L_{\text{Chamfer}}$ and $\mathcal L_{\text{Occ}}$, respectively.
\subsubsection{Performance on Multi-Center CT and MRI Datasets }
Table~\ref{tab:whs_full} summarizes the quantitative performance of different loss functions on the WHS++ CT and MR test datasets. On CT, the Chamfer loss achieves the highest Dice scores across LV, RV, and Myo; however, its CD and HD values are consistently worse than those obtained with the proposed $\mathcal{L}_{\text{MIE}}$, particularly at the summed multi-structure level. In contrast, $\mathcal{L}_{\text{MIE}}$ yields statistically significant reductions in surface distance errors (CD, HD) relative to $\mathcal{L}_{\text{Chamfer}}$, and substantially lower volumetric violations (VR/SVR) than both $\mathcal{L}_{\text{Chamfer}}$ and $\mathcal{L}_{\text{Occ}}$ (p < 0.05), indicating more faithful surface geometry and improved anatomical coherence, albeit with a marginal but acceptable Dice decrease. The plain occupancy variants ($\mathcal{L}_{\text{Occ}}, \mathcal{L}_{\text{Occ}}^{0.2}$) exhibit higher variance and inferior geometric fidelity, particularly under edge-biased resampling. The degradation observed in $\mathcal{L}_{\text{Occ}}^{0.2}$ confirms that naive boundary-focused sampling amplifies local noise without leveraging structured anatomical priors.

On MR, the superiority of $\mathcal{L}_{\text{MIE}}$ becomes more pronounced. It achieves statistically significant Dice gains across all structures and consistently improves CD and/or HD relative to $\mathcal{L}_{\text{Chamfer}}$, with additional advantages over $\mathcal{L}_{\text{Occ}}$ in most cases (p < 0.05), while $\mathcal{L}_{\text{MIE}}^{0.2}$ ranks closely second. Although edge-aware resampling further suppresses local volumetric leakage (as reflected by lower VR/SVR), it also introduces slightly higher inter-structure variance and less stable global balance. This suggests that excessive emphasis on boundary samples biases the optimization toward local edges rather than global anatomical regularity. As a result, $\mathcal{L}_{\text{MIE}}$ provides a more consistent trade-off between geometric precision and anatomical coherence, and is therefore adopted as the default loss configuration in subsequent experiments.

A different trend is observed for the LSE. As defined in Eq.~(\ref{eq:lap}), LSE measures local surface oscillation by quantifying the deviation of each vertex from a Laplacian-smoothed estimate of its neighbors. Lower LSE therefore indicates smoother surface curvature. On both CT and MR, Chamfer supervision consistently yields the lowest LSE values, and the differences are statistically significant compared with $\mathcal{L}_{\text{MIE}}$ and $\mathcal{L}_{\text{MIE}}^{0.2}$ (p < 0.05), indicating that Chamfer favors locally smoother surface adjustments. However, this aggressive smoothing coexists with significantly higher VR/SVR and frequent membrane penetration, suggesting that overly smooth surfaces may be anatomically invalid. In contrast, although $\mathcal{L}_{\text{MIE}}$ exhibits significantly higher LSE values, it achieves statistically significant improvements in HD, CD, VR, and SVR (p < 0.05). This indicates that the proposed loss intentionally relaxes excessive curvature minimization in order to preserve inter-chamber separability and prevent structural interpenetration. In this sense, the increased LSE reflects a physically meaningful trade-off rather than degraded surface quality. Moreover, the distribution and spatial pattern of the average LSE on the mean heart shapes of the WHS++ CT and MR test sets under $\mathcal{L}_{\text{MIE}}$ (Fig.~\ref{lse}) further supports this interpretation. As visualized, higher LSE values are predominantly concentrated along anatomically meaningful boundaries and high-curvature regions (e.g., chamber junctions and septal interfaces), appearing as brighter bands near the edges, while the interior surfaces remain relatively smooth. This confirms that the LSE increase introduced by $\mathcal{L}_{\text{MIE}}$ is spatially structured and anatomically localized, rather than arising from random surface oscillations.

\begin{figure*}[!t] \centerline{\includegraphics[width=\columnwidth]{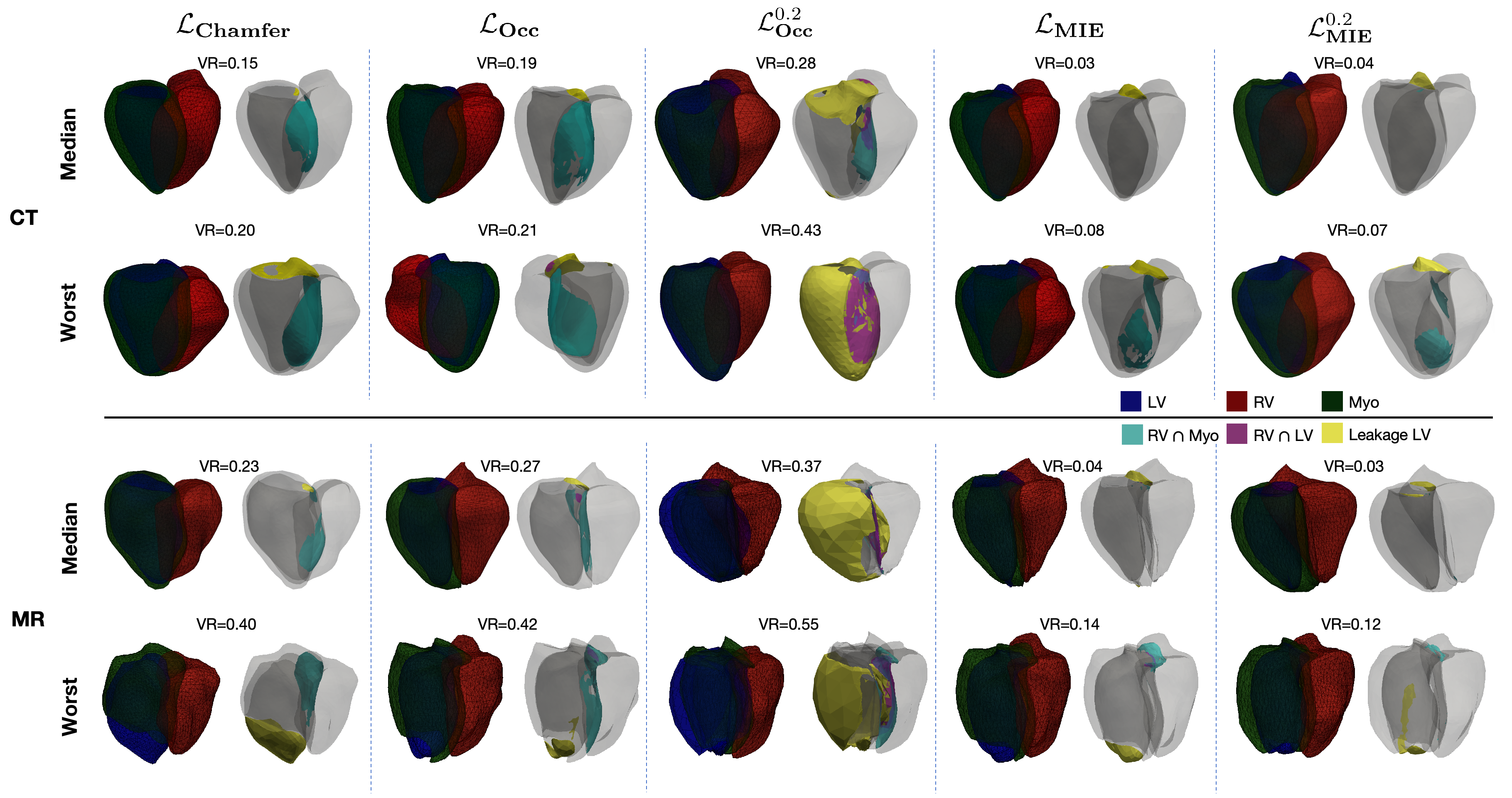}} \caption{The median and worst reconstruction results measured by VR as a scoring metric are visually presented for all comparison methods within the WHS++ test dataset. The three anatomical structures are shown in red (RV), green (Myo), and blue (LV). Intersection regions are highlighted in cyan (RV $\cap$ Myo), magenta (RV $\cap$ LV). The yellow region indicates the leakage volume, representing the portion of LV protruding outside Myo.} 
\label{fig4} 
\end{figure*}

Fig.~\ref{fig4} offers structural insight into these numerical differences. Chamfer- and Occ-based supervision frequently leads to membrane penetration and cross-chamber leakage, violating the anatomical separability of ventricular cavities, despite their occasionally high Dice scores. Such penetrations violate the physical separability of ventricular cavities, undermining their suitability for downstream biomechanical or simulation tasks. In contrast, both $\mathcal L_{\text{MIE}}$ and $\mathcal L_{\text{MIE}}^{0.2}$ eliminate membrane intrusion and preserve myocardial boundaries. The myocardium wall remains continuous and free of perforation, demonstrating that our loss explicitly constrains the reconstruction toward anatomically permissible solutions rather than relying solely on pointwise surface proximity. Between them, $\mathcal L_{\text{MIE}}^{0.2}$ reduces local errors but remains less balanced than $\mathcal L_{\text{MIE}}$, which achieves the most reliable global reconstruction.

\begin{figure*}[!t] \centerline{\includegraphics[width=\columnwidth]{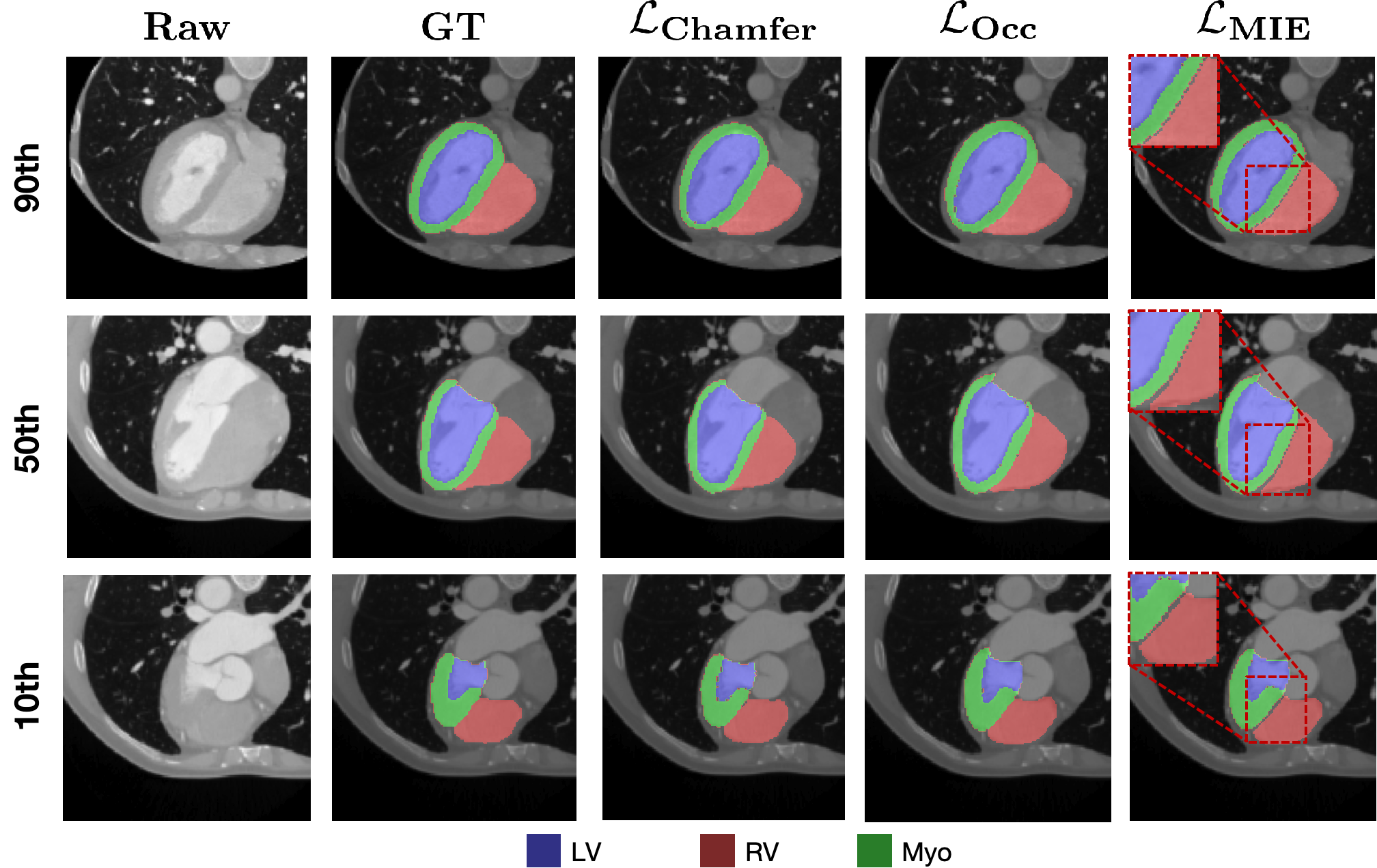}} \caption{Two-dimensional comparison of three cardiac structures in CT test cases predicted by different methods. Each row displays magnified axial image views of test cases alongside their predicted results, annotated with the 10th, 50th, and 90th percentile Dice scores calculated using this method. The proposed $\mathcal L_{\text{MIE}}$ preserves clear anatomical separation between adjacent chambers, as highlighted by the red dashed boxes. LV, RV, and Myo are shown in blue, red, and green, respectively} 
\label{fig5} 
\end{figure*}

\begin{figure*}[!t] \centerline{\includegraphics[width=\columnwidth]{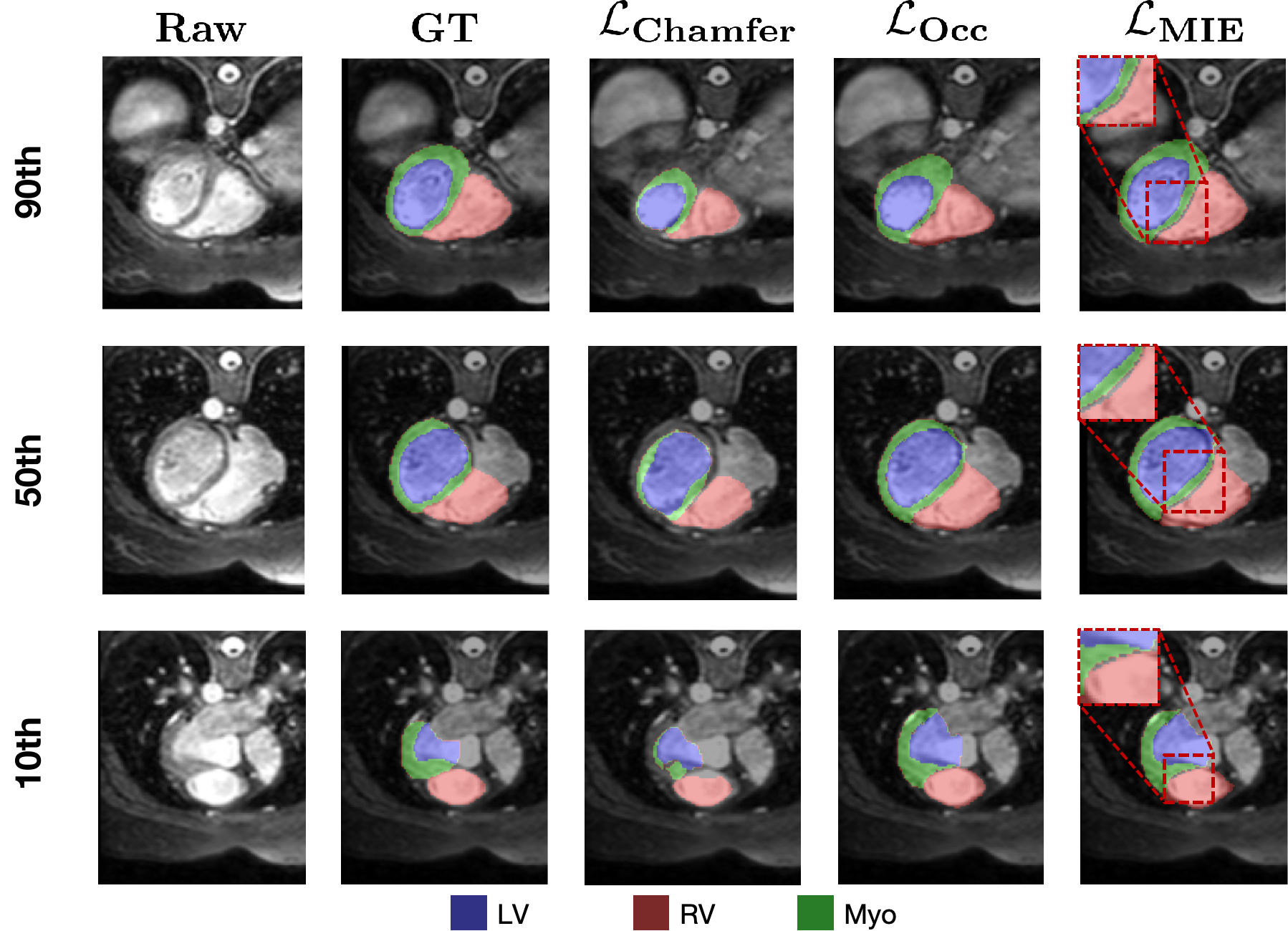}} \caption{Two-dimensional comparison of three cardiac structures in MR test cases predicted by different methods. Each row displays magnified axial image views of test cases alongside their predicted results, annotated with the 10th, 50th, and 90th percentile Dice scores calculated using this method. The proposed $\mathcal L_{\text{MIE}}$ preserves clear anatomical separation between adjacent chambers, as highlighted by the red dashed boxes. LV, RV, and Myo are shown in blue, red, and green, respectively} 
\label{fig6} 
\end{figure*}

\begin{figure*}[!t] \centerline{\includegraphics[width=\columnwidth]{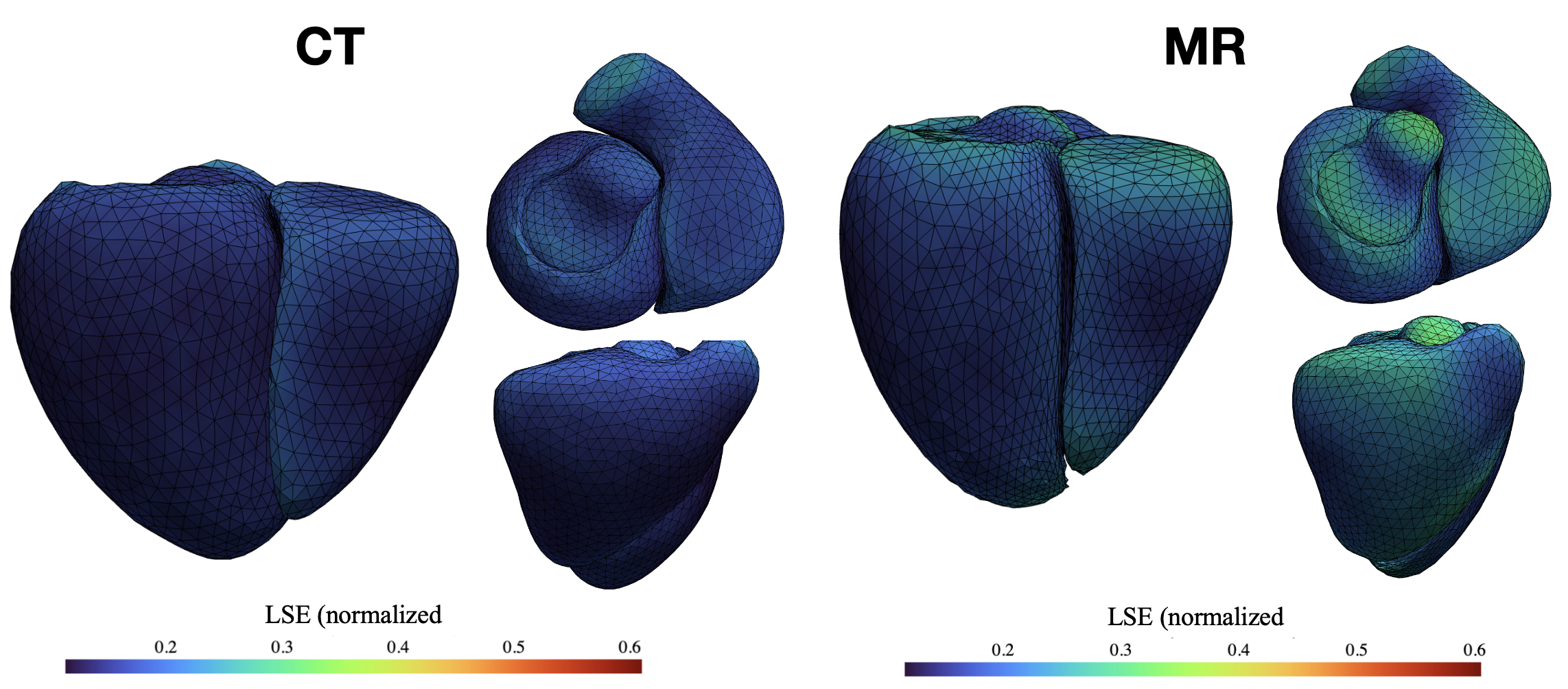}} \caption{Spatial visualization of the normalized average LSE on the mean heart shapes reconstructed with $\mathcal{L}_{\text{MIE}}$ on the WHS++ CT and MR test sets.}
\label{lse} 
\end{figure*}

Fig.~\ref{fig5} and Fig.~\ref{fig6} further confirm this behavior from a slice-wise perspective. Although our loss is optimized in 3D space, the 2D Dice distributions at the 10th, 50th, and 90th percentiles remain competitive on both CT and MR. This demonstrates that the structural reliability observed in 3D is preserved in two-dimensional projections, affirming the consistency of the reconstructed anatomy rather than being driven by isolated regions.

\subsubsection{External Inference on Independent Cohorts}

\begin{figure*}[!t] \centerline{\includegraphics[width=\columnwidth]{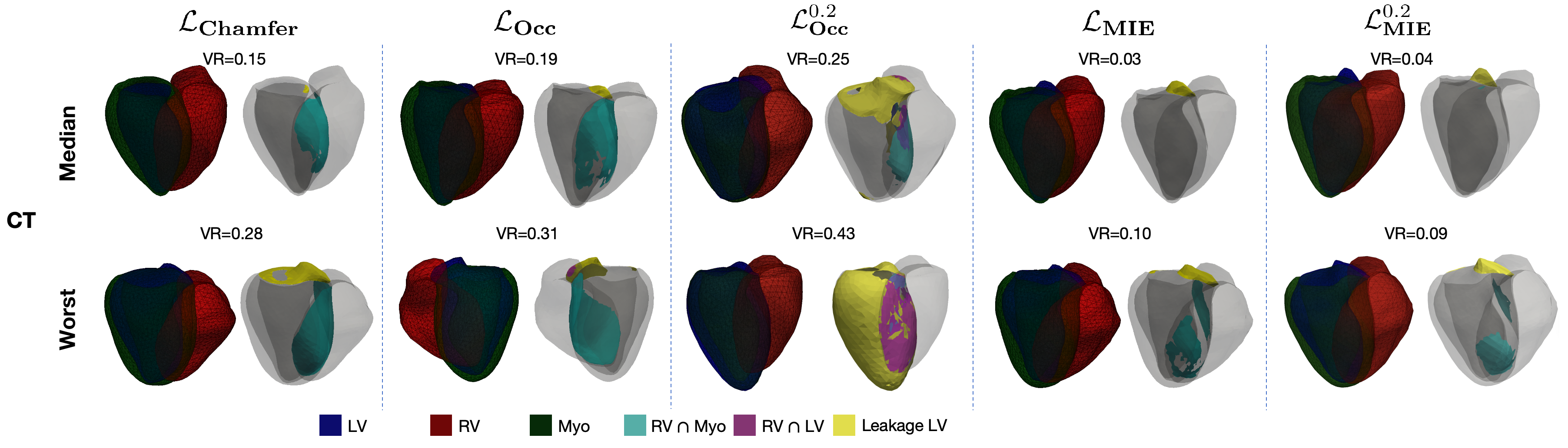}} \caption{The median and worst reconstruction results measured by VR as a scoring metric are visually presented for all comparison methods within the ImageCAS test dataset. The three anatomical structures are shown in red (RV), green (Myo), and blue (LV). Intersection regions are highlighted in cyan (RV $\cap$ Myo), magenta (RV $\cap$ LV). The yellow region indicates the leakage volume, representing the portion of LV protruding outside Myo.} 
\label{fig7} 
\end{figure*}

\begin{figure*}[!t] \centerline{\includegraphics[width=\columnwidth]{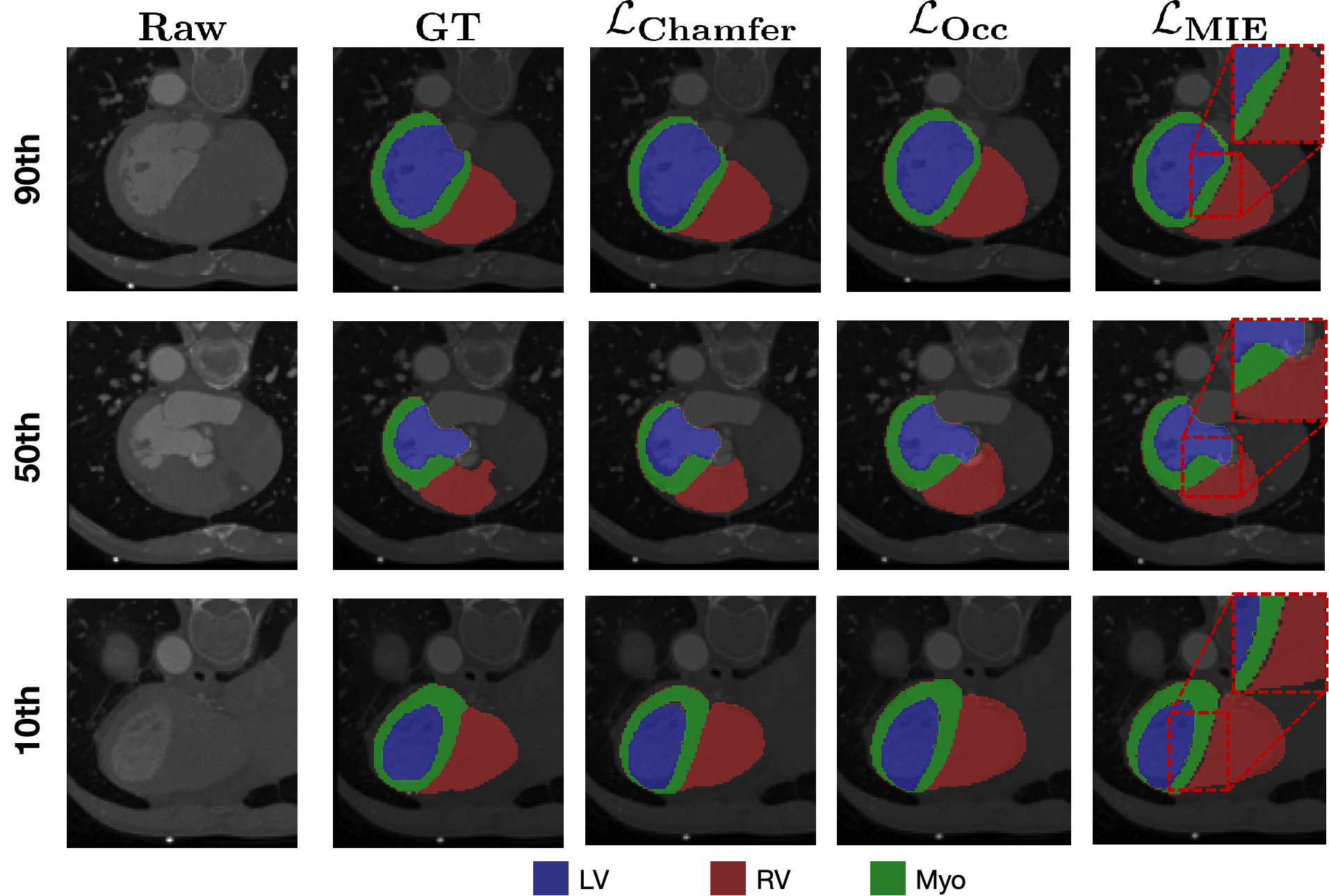}} \caption{Two-dimensional comparison of three cardiac structures in ImageCAS test cases predicted by different methods. Each row displays magnified axial image views of test cases alongside their predicted results, annotated with the 10th, 50th, and 90th percentile Dice scores calculated using this method. The proposed $\mathcal L_{\text{MIE}}$ preserves clear anatomical separation between adjacent chambers, as highlighted by the red dashed boxes. LV, RV, and Myo are shown in blue, red, and green, respectively} 
\label{fig8} 
\end{figure*}

The cross-center evaluation on the ImageCAS CT dataset (Table~\ref{tab:imagecas}, Fig.~\ref{fig7}, and Fig.~\ref{fig8}) demonstrates that the proposed $\mathcal L_{\text{MIE}}$ generalizes reliably beyond the training domain. This cohort differs substantially from WHS++ in terms of scanner type, acquisition protocol, and population characteristics. Although $\mathcal L_{\text{Chamfer}}$ achieves the highest Dice scores on LV and RV, it also produces significantly higher VR/SVR, indicating that surface-distance supervision alone does not effectively prevent cross-chamber leakage under domain shift. In contrast, $\mathcal L_{\text{MIE}}$ preserves competitive Dice while yielding consistently lower VR/SVR across all structures, confirming that its anatomical constraints remain effective under unseen imaging conditions. Qualitative comparisons in Fig.~\ref{fig7} further show that MIE suppresses membrane penetration even in adverse cases, and the percentile slice analysis in Fig.~\ref{fig8} confirms its superior slice-wise consistency. These results indicate that enforcing anatomical relations provides more reliable multi-center generalization than distance-based supervision.

\begin{figure*}[!t] 
\centering
\centerline{\includegraphics[width=\columnwidth]{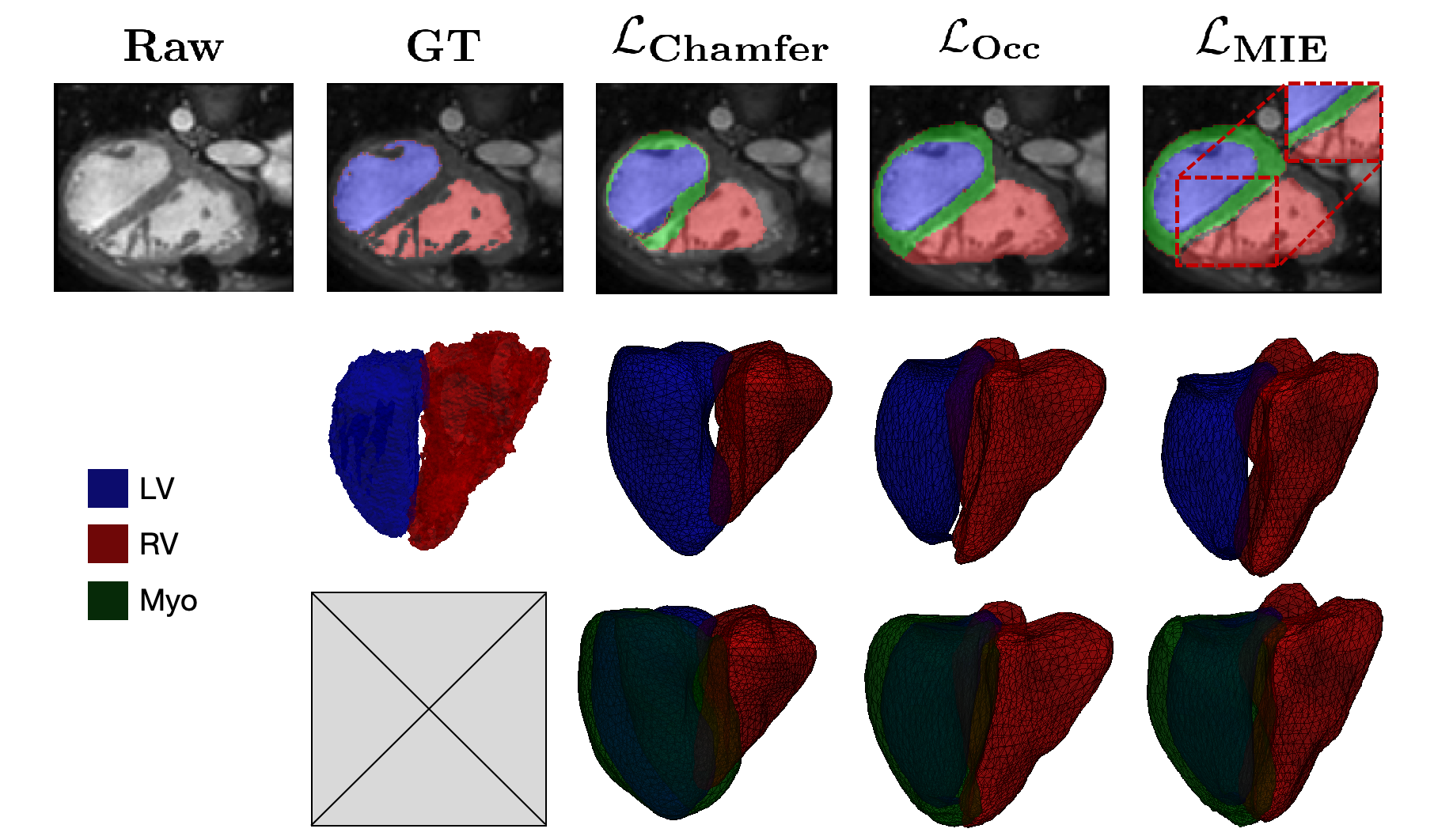}} \caption{
Qualitative comparison of multi-structures reconstruction on a representative HVSMR case. Top row: raw axial image, ground-truth annotations, and predictions 2D slice from different losses. Middle row: corresponding reconstructed 3D meshes for LV (blue) and RV (red). Bottom row: corresponding reconstructed 3D meshes for LV (blue), Myo (Green) and RV (red). The proposed $\mathcal L_{\text{MIE}}$ preserves clear anatomical separation between adjacent chambers, as highlighted by the red dashed boxes.}
\label{fig9} 
\end{figure*}

We further evaluated the proposed framework on the HVSMR dataset, which consists exclusively of bi-ventricular congenital heart disease (CHD) phenotypes with well-defined LV/RV identity. True single-ventricle physiology, for which a consistent LV/RV labeling does not exist, is explicitly outside the scope of the present multi-chamber reconstruction formulation. As illustrated in Fig.~\ref{fig9}, CHD subjects exhibit pronounced ventricular deformation, marked asymmetry, and highly irregular inter-ventricular adjacency, deviating substantially from the spatial regularities that characterize healthy cardiac morphology and making this cohort a stringent test of cross-shape, cross-center, and cross-pathology generalization.

In the 2D slices and corresponding 3D meshes, the model trained with $\mathcal L_{\text{Chamfer}}$ produces ventricles that are overly smoothed and biased toward rounded shapes, failing to reproduce the sharp asymmetry and local contour variations observed in the ground truth. $\mathcal L_{\text{Occ}}$ achieves better volumetric overlap, but the global geometry remains distorted: the curvature of the LV, the extent of RV bulging, and the narrow inter-ventricular gap are not faithfully recovered, leading to a noticeably altered overall heart shape. By contrast, the proposed $\mathcal L_{\text{MIE}}$ yields reconstructions that more faithfully reflect the morphological trends in the ground truth, preserving chamber-specific elongation, orientation, and fine-scale inter-ventricular spacing despite severe pathological distortion.

The quantitative results on the HVSMR cohort in Table~\ref{tab:hvsmr} are consistent with these qualitative findings. $\mathcal L_{\text{Occ}}$ attains the highest Dice scores for both LV and RV, reflecting its strong bias toward volumetric consistency. However, on this CHD dataset—where ventricular geometry and adjacency deviate substantially from normal patterns—high overlap alone is insufficient to guarantee anatomically faithful meshes, as evidenced by the larger surface and Laplacian errors. The proposed $\mathcal L_{\text{MIE}}$ trades a small amount of Dice overlap for a more favorable balance of CD and HD, while accepting higher LSE values. Consistent with our WHS++ analysis, this increase in LSE does not correspond to random surface roughness, but rather to sharper, anatomically meaningful curvature along highly deformed ventricular boundaries and inter-ventricular interfaces. In this sense, $\mathcal L_{\text{MIE}}$ prioritizes structural plausibility and inter-chamber separability over aggressive curvature smoothing, leading to improved global shape fidelity despite the severe morphological distortion in CHD.

Together, these cross-center and cross-pathology evaluations demonstrate that enforcing multi-structure anatomical relations via $\mathcal L_{\text{MIE}}$ provides more robust geometric generalization than conventional distance-based or purely volumetric supervision, supporting its potential for reliable deployment in heterogeneous clinical scenarios.

\begin{table*}[t]
\centering
\scriptsize
\setlength{\tabcolsep}{4pt}
\caption{Performance on the ImageCAS CT sets.
         Higher\,=\,better for DSC; lower\,=\,better for CD/HD/LSE/VR/SVR. $^{*}$ p < 0.05 vs $\mathcal L_{\text{Chamfer}}$;\;$^{\dagger}$ p < 0.05 vs $\mathcal L_{\text{Occ}}$.}
\label{tab:imagecas}
\begin{tabular}{llcccccc}
\toprule
\textbf{Struct.} & \textbf{Method} &
\textbf{DSC} $\uparrow$ & \textbf{CD (mm)} $\downarrow$ &
\textbf{HD (mm)} $\downarrow$ & \textbf{LSE} $\downarrow$ &
\textbf{VR} $\downarrow$ & \textbf{SVR} $\downarrow$\\
\midrule
\multirow{5}{*}{LV}
 & $\mathcal L_{\text{Chamfer}}$               & \textbf{0.9566±0.0135} & 0.0003±0.0001 & \textbf{0.0618±0.0255} & 0.2953±0.0238 & 0.1401±0.0254 & 0.1957±0.0015\\
 & $\mathcal L_{\text{Occ}}$                   & 0.9548±0.0105 & 0.0004±0.0003 & 0.0674±0.0195 & 0.3064±0.0215 & 0.0995±0.0014 & 0.1981±0.0006\\
 & $\mathcal L_{\text{Occ}}^{0.2}$     & 0.5977±0.0297  & 0.0178±0.0044 & 0.1517±0.0169           & 0.3029±0.0231            & 0.3692±0.1617      & 0.2618±0.0148\\
 & $\mathcal L_{\text{MIE}}$                   & 0.9561±0.0135 & \textbf{0.0002±0.0012}& 0.1319±0.0432$^{*}$ & \textbf{0.2719±0.0185}$^{*}$ & 0.0995±0.0008$^{*}$ &0.1941±0.0014$^{*}$\\
 & $\mathcal L_{\text{MIE}}^{0.2}$     & 0.9480±0.0095  & 0.0004±0.0001 & 0.0667±0.0226            & 0.3059±0.0214       &\textbf{ 0.0609± 0.0016}   & \textbf{0.1746±0.0013}\\
\cmidrule(lr){2-8}
\multirow{5}{*}{RV}
 & $\mathcal L_{\text{Chamfer}}$               & \textbf{0.8968±0.0347} & 0.0020±0.0017& 0.1264±0.0452 & \textbf{0.2563±0.0196} & 0.0921±0.0247 & 0.2266±0.0043\\
 & $\mathcal L_{\text{Occ}}$                   & 0.8567±0.0331$^{*}$  & 0.0024±0.0016$^{*}$ & 0.1348±0.0507$^{*}$ & 0.2704±0.0186$^{*}$ & 0.1250±0.0197 & 0.2262±0.0022\\
 & $\mathcal L_{\text{Occ}}^{0.2}$     & 0.8701±0.0353  & \textbf{0.0017±0.0015}  & 0.1302±0.0471 & 0.2899±0.0221 & 0.2910±0.0562 & 0.2281±0.0038\\
 & $\mathcal L_{\text{MIE}}$                   & 0.8744±0.0292$^{*}$$^{\dagger}$ & 0.0018±0.0009$^{*}$$^{\dagger}$ & \textbf{0.1166±0.0286}$^{*}$$^{\dagger}$ & 0.2641±0.0271$^{*}$$^{\dagger}$ &  0.0307±0.0174$^{*}$$^{\dagger}$ &0.2015±0.0132$^{*}$$^{\dagger}$\\
 & $\mathcal L_{\text{MIE}}^{0.2}$     & 0.8609±0.330 & 0.0019±0.0011  & 0.1304±0.0408            & 0.2697±0.0181 &  \textbf{0.0224±0.0148}          &  \textbf{0.1982±0.0131}\\
\cmidrule(lr){2-8}
\multirow{5}{*}{Myo}
 & $\mathcal L_{\text{Chamfer}}$               & \textbf{0.8700±0.0329} & 0.0005±0.0006 & 0.0712±0.0219 & \textbf{0.2298±0.0182} & 0.4573±0.0240 & 0.3281±0.0013\\
 & $\mathcal L_{\text{Occ}}$                   & 0.8525±0.0033$^{*}$ & 0.0007±0.0005 & 0.0801±0.0216 & 0.2401±0.0163 & 0.4460±0.0209 & 0.2257±0.0018\\
 & $\mathcal L_{\text{Occ}}^{0.2}$     & -  & 0.0005±0.0002            & \textbf{0.0711±0.0242}            & 0.2582±0.0192          & 0.5971±0.0605 & 0.2418±0.0206\\
 & $\mathcal L_{\text{MIE}}$                   & 0.8521±0.0397$^{*}$ & 0.0004±0.0014 & 0.1001±0.0435$^{*}$$^{\dagger}$ & 0.2249±0.0198$^{\dagger}$ & 0.1001±0.0066$^{*}$$^{\dagger}$ & 0.2279±0016$^{*}$$^{\dagger}$\\
 & $\mathcal L_{\text{MIE}}^{0.2}$     & 0.8467±0.0321 & \textbf{0.0005±0.0001} & 0.0781±0.0203            &0.2282±0.0156 & \textbf{0.0994±0.0046 }& \textbf{0.1978±0.0014}\\
\cmidrule(lr){2-8}
\multirow{5}{*}{Sum}
 & $\mathcal L_{\text{Chamfer}}$               & \textbf{0.9008±0.0217} & 0.0028±0.0014 & \textbf{0.2595±0.0745} & \textbf{0.7815±0.0613} & 0.1564±0.0662 & 0.2820±0.0016\\
 & $\mathcal L_{\text{Occ}}$                   & 0.8886±0.0222$^{*}$ & 0.0036±0.0017 & 0.2898±0.0736$^{*}$ & 0.8141±0.0558$^{*}$ & 0.1926±0.0673 & 0.2803±0.0012\\
 & $\mathcal L_{\text{Occ}}^{0.2}$     & 0.4915±0.0126 & 0.0204±0.0045 & 0.3531±0.0664            & 0.8511±0.0642           & 0.3589±0.0541     & 0.2910±0.0086\\
 & $\mathcal L_{\text{MIE}}$                   & 0.8991±0.0257 & \textbf{0.0024±0.0015}$^{\dagger}$ & 0.2776±0.0681$^{*}$ & 0.8114±0.0561$^{*}$ & 0.0695±0.0050$^{*}$$^{\dagger}$ & 0.1209±0.0052$^{*}$$^{\dagger}$\\
 & $\mathcal L_{\text{MIE}}^{0.2}$     & 0.8897±0.0208 & 0.0029±0.0012 & 0.2752±0.0654            &0.8039±0.0546            & \textbf{0.0422±0.0059 }    & \textbf{0.1096±0.0051}\\
\bottomrule
\end{tabular}
\end{table*}

\section{Discussion}
The results in Table~\ref{tab:whs_full} and Table~\ref{tab:imagecas} demonstrate that merely improving overlap metrics does not guarantee anatomically trustworthy cardiac surface reconstruction. On CT, Chamfer supervision produces the highest Dice scores across all structures, yet it also leads to membrane penetration and cross-chamber leakage. By contrast, the proposed MIE-based losses yield more stable geometry on both CT and MR, and achieve the best overall performance on MR where image contrast is weaker. This indicates that enforcing inter-structural constraints contributes more directly to anatomical plausibility than minimizing point-wise surface distances alone.
\begin{table}[t]
\centering
\scriptsize
\setlength{\tabcolsep}{1pt}
\caption{Performance on the HVSMR sets.
         Higher\,=\,better for DSC; lower\,=\,better for CD/HD/LSE. $^{*}$ p < 0.05 vs $\mathcal L_{\text{Chamfer}}$;\;$^{\dagger}$ p < 0.05 vs $\mathcal L_{\text{Occ}}$.}
\label{tab:hvsmr}
\begin{tabular}{llcccc}
\toprule
\textbf{Struct.} & \textbf{Method} &
\textbf{DSC} $\uparrow$ & \textbf{CD (mm)} $\downarrow$ &
\textbf{HD (mm)} $\downarrow$ & \textbf{LSE} $\downarrow$ \\
\midrule
\multirow{3}{*}{LV}
 & $\mathcal L_{\text{Chamfer}}$               & 0.7069±0.1120 & 0.0074±0.0045 & 0.2164±0.0748 & \textbf{0.2433±0.0211} \\
 & $\mathcal L_{\text{Occ}}$                   & \textbf{0.8050±0.1405}$^{*}$ & 0.0066±0.0046$^{*}$ & \textbf{0.2142±0.0703} & 0.3259±0.0340$^{*}$ \\
 & $\mathcal L_{\text{MIE}}$                   & 0.7945±0.1479$^{*}$ & \textbf{0.0061±0.0039}$^{*}$$^{\dagger}$ & 0.2176±0.0741 & 0.3344±0.0346$^{*}$ \\
\cmidrule(lr){2-6}
\multirow{3}{*}{RV}
 & $\mathcal L_{\text{Chamfer}}$               & 0.4924±0.1884 & 0.0203±0.0175& 0.3226±0.1337 & \textbf{0.2118±0.0186} \\
 & $\mathcal L_{\text{Occ}}$                   & \textbf{0.7034±0.1998}$^{*}$  & 0.0083±0.0072$^{*}$ & 0.2210±0.0732$^{*}$ & 0.2963±0.0348$^{*}$ \\
 & $\mathcal L_{\text{MIE}}$                   & 0.6906 ± 0.1907$^{*}$ & \textbf{0.0082±0.0080}$^{*}$ & \textbf{0.2108±0.0672}$^{*}$ & 0.3058±0.0329$^{*}$ \\
\cmidrule(lr){2-6}
\multirow{3}{*}{Sum}
 & $\mathcal L_{\text{Chamfer}}$               & 0.5996±0.1237$^{*}$ & 0.0277±0.0202 & 0.5390±0.1836 & \textbf{0.4551±0.0389} \\
 & $\mathcal L_{\text{Occ}}$                   & \textbf{0.7542±0.1560}$^{*}$ & 0.0149±0.0114$^{*}$ & 0.4351±0.1239$^{*}$ & 0.6222±0.0684$^{*}$ \\
 & $\mathcal L_{\text{MIE}}$                   & 0.7425 ± 0.1548$^{*}$ & \textbf{0.0143±0.0127}$^{*}$$^{\dagger}$ & \textbf{0.4285±0.1236}$^{*}$ & 0.6402±0.0672$^{*}$\\

\bottomrule
\end{tabular}
\end{table}

A key observation from Table~\ref{tab:whs_full} is the divergence between Dice and geometry-aware metrics. Dice measures volumetric overlap but does not penalize boundary violations such as wall-crossing, explaining why Chamfer may score well on CT despite structurally invalid outputs. Conversely, HD, CD jointly reflect boundary fidelity, and MIE-based supervision consistently achieves lower errors on these measures. These findings suggest that multi-chamber cardiac reconstruction should not be evaluated by Dice alone, and that spatial-aware metrics are essential for assessing clinical usability.

Beyond Dice and surface-distance metrics, the behavior of VR and SVR further supports this interpretation. As defined in Eq.\ref{eq:vr}–\ref{eq:svr}, VR quantifies the proportion of critical points that violate an inclusion or exclusion constraint, whereas SVR accumulates the violation margin. Because both metrics are relation-normalized, their denominators change with the number of anatomical constraints associated with each structure. As a result, single-structure VR/SVR may appear small, while the summed VR/SVR increases when additional relations contribute to the normalization term. This explains the discrepancy between chamber-wise and whole-heart scores in Table~\ref{tab:whs_full}, and implies that VR/SVR should be interpreted relationally rather than independently. Such behavior arises intrinsically from the metric formulation rather than reflecting instability of the reconstruction. Importantly, MIE and MIE-edge Losses still yield the lowest VR and SVR values among all losses, demonstrating that the proposed supervision effectively enforces anatomical separability.

Differences between CT and MR performance can also be explained by modality characteristics. CT offers sharper boundaries, allowing Chamfer to fit contours effectively, but its lack of relational awareness leads to structural violations. MR, on the other hand, exhibits greater ambiguity in chamber borders, amplifying the limitations of surface-only supervision and favoring MIE Loss, whose priors do not rely on strong boundary contrast. These modality-specific trends indicate that structural priors reduce dependence on local image intensity and improve robustness under challenging conditions.

The behavior of LSE reflects an expected trade-off introduced by enforcing anatomical relations. Since the Laplacian energy penalizes local curvature (Eq.\ref{eq:lap}), overly smooth surfaces minimize LSE, but may oversuppress physiologically meaningful transitions. In contrast, $\mathcal{L}_{\mathrm{MIE}}$ preserves inter-chamber boundaries by allowing controlled local deformation, which can moderately increase LSE. We consider this a reasonable side-effect rather than a limitation, as anatomical validity is prioritized over purely geometric smoothness, particularly in datasets where inter-chamber constraint is well defined. Under extreme pathological deformation (e.g., HVSMR), LSE may further increase as a result of highly irregular curvature patterns, which reflects the intrinsic complexity of the anatomy rather than instability of the reconstruction. Future work may combine relational supervision with anatomy-aware smoothing priors to further balance these factors.

Although the proposed model is trained exclusively on anatomically normal multi-center datasets (MMWHS and WHS++), its behavior on external cohorts reveals important properties of the MIE regularization. On the ImageCAS CT dataset, which differs substantially in scanner characteristics, contrast profile, and reconstruction kernel, $\mathcal{L}_{\mathrm{MIE}}$ maintains stable surface accuracy and structure separation, indicating strong robustness to center-specific imaging variations. More importantly, the evaluation on the HVSMR congenital heart disease cohort highlights the model’s capacity to operate under severe morphological deviations. Because CHD cases violate key spatial regularities learned from healthy hearts—such as chamber adjacency, symmetry, and global proportions—distance-only losses such as $\mathcal{L}_{\mathrm{Chamfer}}$ or $\mathcal{L}_{\mathrm{Occ}}$ tend to produce implausible geometry or irregular ventricular boundaries. In contrast, $\mathcal{L}_{\mathrm{MIE}}$ shows more stable multi-structure relationships even when the underlying anatomy departs dramatically from normal patterns. This indicates that the relational priors encoded by $\mathcal{L}_{\mathrm{MIE}}$ confer a form of robustness to pathological variability, enabling stable reconstruction despite shape variability and cross-population distribution shifts. Such findings support the potential applicability of the framework to anatomically diverse or clinically challenging scenarios where explicit multi-structure annotation is limited or unavailable.

Beyond numerical improvements, the proposed framework highlights a broader conceptual shift: multi-structure cardiac reconstruction should be governed by anatomical relations in addition to geometric proximity. This is particularly relevant in clinical scenarios where manual 3D annotation is scarce and inter-structure consistency is assessed implicitly by clinicians through spatial reasoning rather than explicit volumetric measures. By bridging this gap, relational supervision provides a mechanism to inject clinically meaningful structural knowledge directly into geometric learning.

Beyond cardiac reconstruction, the proposed relational supervision framework is conceptually applicable to other multi-organ or multi-compartment anatomical systems in which spatial relations govern structural plausibility, such as liver–vessel complexes, pelvic organs, or craniofacial structures. While the specific relational rules would need to be redesigned for each anatomical context, this suggests that relational mesh supervision has the potential to inspire a broader class of anatomy-aware shape reconstruction methods beyond the cardiac domain.

Several limitations suggest natural extensions of this work. First, the current formulation enforces anatomical relations at the surface level, without explicitly modeling wall thickness, myocardial fiber orientation, or valve geometry. These structures are physiologically meaningful and may serve as additional constraints to further tighten anatomical plausibility. Second, despite the demonstrated robustness across multiple CT centers, our MR validation remains constrained by data availability. Dense and fully-sampled MR volumes are difficult to acquire in routine clinical practice, and the scarcity of such datasets limits our ability to systematically assess the proposed relational loss under diverse MR imaging conditions. Future work will focus on evaluating the framework on accelerated or sparsely sampled MR protocols, and on incorporating MR-specific anatomical priors to further extend applicability in realistic clinical settings. Finally, the heart is intrinsically four-dimensional, whose structural relationships evolve continuously throughout the cardiac cycle. The current framework enforces anatomical consistency only in static meshes, without constraining temporal coherence or motion-dependent chamber interactions. Extending relational supervision to 4D cine reconstruction, where anatomy and motion are jointly preserved, would therefore represent a natural completion of the proposed formulation. Such an extension may further improve physiological plausibility, particularly in regions where functional coupling rather than static geometry determines anatomical boundaries.

\section{Conclusions}

We present a relational anatomical supervision framework for multi-chamber cardiac mesh reconstruction that explicitly embeds inter-structural consistency into differentiable geometric learning. Through extensive evaluation on multi-center CT data, densely sampled MR data, and two independent external cohorts including congenital heart disease, we demonstrate that conventional overlap- and distance-based supervision is insufficient to guarantee anatomically valid geometry, whereas the proposed MIE loss robustly suppresses boundary violations and preserves physiologically meaningful chamber organization.

Beyond improving numerical accuracy, this work establishes that anatomically faithful cardiac reconstruction is fundamentally a relational problem rather than a purely geometric one. By jointly introducing relation-aware supervision and violation-centric evaluation, we provide both a modeling and an assessment paradigm for structure-preserving cardiac geometry learning. The consistent gains observed under severe domain shifts and pathological deformation further indicate that relational anatomical priors offer a powerful and generalizable form of regularization.

More broadly, the proposed framework offers a foundation for anatomy-aware geometric learning beyond static surface reconstruction. It can be naturally extended to incorporate additional cardiac substructures, modality-specific priors, and spatiotemporal constraints for 4D cine modeling. We believe that embedding explicit anatomical relations into data-driven reconstruction may represent a critical step toward reliable patient-specific digital heart models, with potential impact on physiological simulation, image-guided intervention, and future data-driven cardiac digital twins.

\section*{Acknowledgments}

Guang Yang was supported in part by the ERC IMI (101005122), the H2020 (952172), the MRC (MC/PC/21013), the Royal Society (IEC\textbackslash NSFC\textbackslash211235), the NVIDIA Academic Hardware Grant Program, the SABER project supported by Boehringer Ingelheim Ltd, NIHR Imperial Biomedical Research Centre (RDA01), The Wellcome Leap Dynamic resilience program (co-funded by Temasek Trust)., UKRI guarantee funding for Horizon Europe MSCA Postdoctoral Fellowships (EP/Z002206/1), UKRI MRC Research Grant, TFS Research Grants (MR/U506710/1), Swiss National Science Foundation (Grant No. 220785), and the UKRI Future Leaders Fellowship (MR/V023799/1, UKRI2738).\\
I sincerely appreciate the support and invaluable help of Dio who assisted in the design of the tables, making the data presentation clearer and more effective.

\bibliographystyle{model2-names.bst}\biboptions{authoryear}
\bibliography{refs}



\end{document}